\documentclass[letterpaper]{article} % DO NOT CHANGE THIS
\usepackage{aaai2027} % Public arXiv preprint: show author information
\nocopyright % Suppress the AAAI proceedings copyright notice in this preprint
\usepackage[hyphens]{url} % DO NOT CHANGE THIS
\usepackage{graphicx} % DO NOT CHANGE THIS
\usepackage{natbib} % DO NOT CHANGE THIS
\usepackage{caption} % DO NOT CHANGE THIS
\usepackage{booktabs}
\usepackage{multirow}
\usepackage{amsmath,amssymb,amsfonts}
\usepackage{threeparttable}
\usepackage{algorithm}
\usepackage{algpseudocode}
\usepackage[capitalize]{cleveref}
\usepackage{xspace}
\usepackage[table]{xcolor}
\usepackage{placeins}
\newcommand{\comment}[1]{}

\newcommand{\pdfurilink}[2]{%
  \leavevmode
  \pdfstartlink
    attr{/Border [0 0 0]}
    user{/Subtype /Link /A << /S /URI /URI (#1) >>}%
  #2%
  \pdfendlink
}

\newcommand{\meanstd}[2]{\ensuremath{#1 \pm #2}}
\newcommand{\meanstdbf}[2]{\ensuremath{\mathbf{#1 \pm #2}}}

\newcommand{\appitem}[2]{%
  \noindent\textbf{\ref{#2}}\hspace{0.6em}\textbf{#1}\dotfill\pageref{#2}\par
}
\newcommand{\appsubitem}[2]{%
  \noindent\hspace*{1.2em}\ref{#2}\hspace{0.6em}#1\dotfill\pageref{#2}\par
}

\begin{document}
\title{Event ActivityNet: A Large-Scale Simulated-Event Benchmark for Untrimmed Action Understanding}

% Public arXiv preprint version with visible author information.
\author{
    Cheng-Yao Hong\textsuperscript{\rm 1},
    Ting-Wei Lin\textsuperscript{\rm 1},
    Yun-Chung Lai\textsuperscript{\rm 2},
    Hua-Wei Lee\textsuperscript{\rm 2},\\
    Hwann-Tzong Chen\textsuperscript{\rm 2},
    Tyng-Luh Liu\textsuperscript{\rm 1}
}
\affiliations{
    \textsuperscript{\rm 1}Institute of Information Science, Academia Sinica, Taiwan\\
    \textsuperscript{\rm 2}Department of Computer Science, National Tsing Hua University, Taiwan
}

\maketitle

\begin{abstract}
Long-horizon event-based action understanding remains underexplored because existing datasets largely comprise short, trimmed clips, while collecting native event streams with dense temporal annotations is costly. We introduce \textbf{Event~ActivityNet}, a large-scale simulated-event benchmark derived from human-annotated, untrimmed ActivityNet videos. It comprises 3{,}263 videos, 200 action classes, and 106.94 hours, with matched 5-bin and 9-bin event-voxel representations, temporal action annotations, and timestamped captions. The benchmark supports annotated-segment action recognition, auxiliary event--language alignment, and causal online temporal action localization. We generate event voxels directly from non-interpolated source videos in decoded frame order, retain per-video rational nominal or average frame-rate metadata for approximate time mapping, and use action-center reconstruction LPIPS as a soft diagnostic of retained reconstructable content. We establish baselines for adaptive event framing, prompt--caption alignment, and event-only, RGB-only, and RGB--event localization. Under a progressive nested-scale training protocol, recognition Top-1 accuracy increases from 52.25 to 66.42, while online temporal localization average mAP improves from 21.7 to 29.0. Moreover, staged Event~ActivityNet pretraining followed by native-event fine-tuning consistently outperforms target-only and joint-from-scratch training across multiple supervision budgets. Event~ActivityNet provides a scalable benchmark for long-horizon event modeling, although native-camera evaluation remains essential for deployment-oriented conclusions.
\end{abstract}

\begin{links}
    \noindent\textbf{Dataset} --- 
    \pdfurilink{https://huggingface.co/datasets/IIS-CVL/EventActivityNet}
    {\texttt{Event ActivityNet on Hugging Face}}
\end{links}

\begin{figure*}[t]
    \centering
    \includegraphics[width=0.97\textwidth]{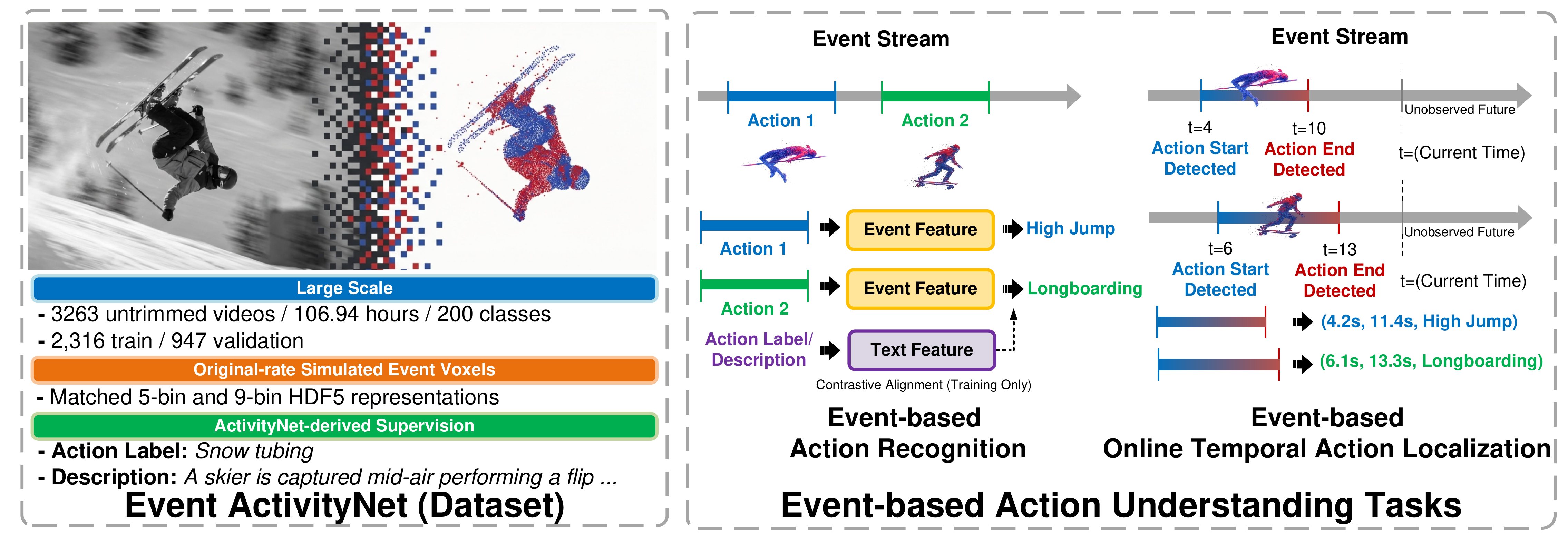}
    \caption{\textbf{Event~ActivityNet at a glance.}
Left: Event~ActivityNet comprises 3{,}263 untrimmed videos (106.94 hours) across 200 action classes, converted directly into simulated event voxels at source-video frame rates. Matched 5-bin and 9-bin HDF5 representations enable controlled temporal-discretization studies. It inherits ActivityNet v1.3 action annotations and ActivityNet Captions descriptions, with tIoU-based caption--action alignments providing auxiliary language supervision. Right: recognition uses annotated action segments and inherited labels, whereas online TAL causally processes untrimmed streams. Text is used only for training-time contrastive alignment, while RGB--event fusion is evaluated separately as a paired-modality baseline.}
    \label{fig:teaser}
\end{figure*}

\begin{table*}[t]
\centering
\caption{\textbf{Comparison with event-based action datasets.}
The listed existing benchmarks use native events and trimmed clips for action recognition. Event ActivityNet instead provides simulated-event representations of untrimmed videos, temporal action annotations, and a causal online TAL protocol.}
\label{tab:dataset_comparison}
\resizebox{\textwidth}{!}{
\begin{tabular}{lccccccc}
\toprule
\textbf{Dataset} & \textbf{Year} & \textbf{Source} & \textbf{Resolution} & \textbf{Classes} & \textbf{Videos / Clips} & \textbf{Temporal Format} & \textbf{Task} \\
\midrule
DVS128-Gesture~\cite{amir2017low} & 2017 & Native & $128 \times 128$ & 11 & 1,342 & Trimmed & Recognition \\
DailyAction~\cite{LiuXTM021} & 2021 & Native & $346 \times 260$ & 12 & 2,000 & Trimmed & Recognition \\
THU$^{E-ACT}$-50~\cite{GaoLLMDLD23} & 2023 & Native & $1280 \times 800$ & 50 & 12,830 & Trimmed & Recognition \\
HARDVS~\cite{wang2024hardvs} & 2024 & Native & $346 \times 260$ & 300 & 100,000+ & Trimmed & Recognition \\
DailyDVS-200~\cite{wang2024daily} & 2024 & Native & $346 \times 260$ & 200 & 22,000 & Trimmed & Recognition \\
THU$^{MV-EACT}$-50~\cite{GaoLLLD24} & 2024 & Native & $1280 \times 800$ & 50 & 31,500 & Trimmed & Recognition \\
CeleX-HAR~\cite{wang2026celexhar} & 2026 & Native & $1280 \times 800$ & 150 & 124,625 & Trimmed & Recognition \\
\midrule
Event ActivityNet & 2026 & Simulated & Various$^*$ & 200 & 3,263 & Untrimmed & Recognition/Online TAL \\
\bottomrule
\end{tabular}
}
\smallskip
{\scriptsize\noindent\hspace*{\tabcolsep}%
\parbox{\dimexpr\textwidth-2\tabcolsep\relax}{\raggedright
$^*$Derived from ActivityNet Captions v1.3~\cite{KrishnaHRFN17}, whose videos originate from ActivityNet v1.3~\cite{caba2015activitynet}; source resolutions range from $128 \times 96$ to $1280 \times 2276$. Event ActivityNet provides precomputed simulated event-voxel representations. Counts follow the units reported by each benchmark; trimmed clip counts and untrimmed video counts are therefore not directly comparable.\par}}
\end{table*}

%=================================================
\section{Introduction}
\label{sec:intro}

Event cameras asynchronously record per-pixel brightness changes, offering low latency, high temporal resolution, and high dynamic range compared with conventional frame cameras~\cite{lichtsteiner2008128,gallego2022event}. These properties benefit high-speed motion, challenging illumination, and streaming perception. Yet long-horizon \emph{action understanding} remains underexplored in the event domain, where existing datasets largely consist of short, trimmed clips with predefined temporal boundaries.

In the RGB domain, ActivityNet~\cite{caba2015activitynet} and THUMOS~\cite{idrees2017thumos} established temporal-localization protocols for untrimmed videos. Online temporal action localization (TAL) further requires predictions at time $t$ to use only observations up to $t$~\cite{KangKKK21,abs-2211-04905}. Comparable large-scale event-stream protocols remain scarce because collecting long native-event recordings with dense temporal boundaries and human-written descriptions is costly. ActivityNet v1.3 provides temporal action annotations, while ActivityNet Captions~\cite{KrishnaHRFN17} supplies timestamped descriptions for the same videos. Together, they enable a simulated-event testbed that reuses existing human supervision without constructing an LLM-generated instruction corpus.

We introduce \textbf{Event~ActivityNet}, a scalable simulated-event testbed for action understanding in untrimmed streams. It supports annotated-segment action recognition and causal online TAL, comprising 3{,}263 videos, 200 action classes, and 106.94 hours. We generate matched 5-bin and 9-bin event-voxel representations directly from non-interpolated source videos, without intermediate 240-Hz videos or continuous raw-event files. Each video retains its nominal or average frame rate, while voxel construction follows decoded frame order rather than per-frame presentation timestamps. Both representations share identical videos, splits, and annotations, enabling controlled studies of temporal discretization.

As shown in Figure~\ref{fig:teaser}, the release integrates event representations, inherited supervision, and reproducible evaluation protocols. It includes a fixed 2{,}316/947 training/validation split, action annotations, timestamped captions, derived caption--action alignments, timing metadata, annotation-issue records, nested-scale manifests, and shard-level integrity information. Our main contributions are:
\begin{itemize}
    \item \textbf{A scalable and auditable simulated-event testbed.}
    Event~ActivityNet extends event-based evaluation beyond trimmed clips to annotated-segment recognition and causal online TAL in untrimmed streams.

    \item \textbf{Reference models for non-stationary event streams.}
    AEF-Split\&Merge adaptively partitions bursty and silent intervals, while a unified contrastive objective aligns event embeddings with class prompts and temporally matched captions without modifying the inherited labels or boundaries.

    \item \textbf{Controlled benchmark analyses.}
    We study progressive nested-scale training, 5-bin versus 9-bin discretization, matched RGB-only, event-only, and paired RGB--event TAL, boundary refinement, and transfer to native event-recognition datasets.
\end{itemize}

Event~ActivityNet does not reproduce sensor-specific noise, native asynchronous timestamps, or high-dynamic-range measurements absent from the source RGB videos. It instead provides a scalable testbed for representation learning, protocol development, and controlled evaluation; native event-camera data remain necessary for deployment-oriented conclusions.

\begin{figure*}[t]
    \centering
    \includegraphics[width=0.97\textwidth]{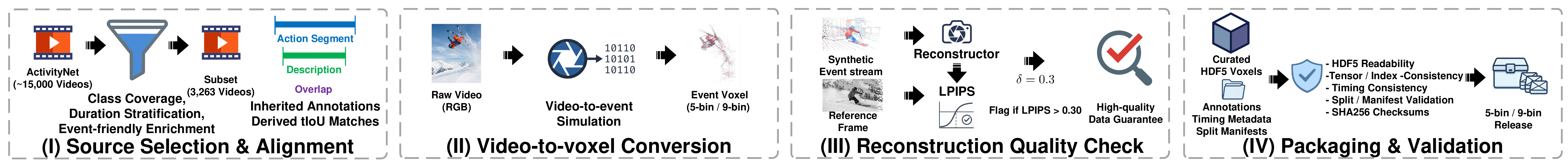}
    \caption{\textbf{Event ActivityNet Construction Pipeline.}
    (1) ActivityNet videos are selected and paired with inherited action annotations and captions, from which fixed splits and caption--action matches are derived.
    (2) Consecutive decoded grayscale frames are converted into dense signed transition slices and grouped into discrete event voxels without high-frame-rate interpolation or intermediate continuous event streams.
    (3) For every annotated action segment, HyperE2VID reconstructs a frame sequence from the voxel window centered at the action midpoint; only the center reconstruction is compared with the nominally aligned source frame using LPIPS.
    (4) The complete finalized payload is packaged and verified through HDF5 readability, tensor/index consistency, timing metadata, split membership, manifests, annotation-issue records, and shard-level checksums.}
    \label{fig:pipeline}
\end{figure*}

%=================================================
\section{Related Work}
\label{sec:related}

\subsubsection{Event Action Datasets.}
Early event-based action recognition focused on trimmed datasets such as DVS128-Gesture~\cite{amir2017low}. Later benchmarks, including THU$^{E-ACT}$-50~\cite{GaoLLMDLD23}, THU$^{MV-EACT}$-50~\cite{GaoLLLD24}, HARDVS~\cite{wang2024hardvs}, DailyDVS-200~\cite{wang2024daily}, and CeleX-HAR~\cite{wang2026celexhar}, substantially increased class diversity and scale, but still emphasize clip-level recognition with predefined boundaries. As shown in~\Cref{tab:dataset_comparison}, Event~ActivityNet instead provides untrimmed simulated-event streams with instance-level temporal annotations and supports causal online localization.

Event streams are commonly encoded as event frames, time surfaces, or voxel grids for use with conventional deep networks. Voxel grids~\cite{zhu2019unsupervised} balance temporal structure and computational efficiency, while scalable video-to-voxel simulation supports larger training collections~\cite{abs-2505-16797}. RGB--event models further combine motion and appearance cues~\cite{GehrigRGHS21}. In long streams, however, fixed windows can mix motion bursts with silent intervals. Prior adaptive accumulation methods~\cite{ZhouZLW24} motivate our voxel-native AEF-Split\&Merge.

%-------------------------------------------------
\subsubsection{Untrimmed and Online TAL.}
Temporal action localization has been widely studied on untrimmed RGB benchmarks such as ActivityNet~\cite{caba2015activitynet} and THUMOS-14~\cite{idrees2017thumos}. Online TAL additionally requires causal inference, as formalized by CAG-QIL~\cite{KangKKK21} and SimOn~\cite{abs-2211-04905}. SimOn queries recent visual and contextual features with the current feature, whereas MATR~\cite{song2024online} uses a selective memory queue and separate start/end decoding paths to capture longer causal context. Comparable protocols remain largely absent from event-action benchmarks, which predominantly assume predefined clip boundaries. Long-form RGB datasets such as Ego4D~\cite{grauman2022ego4d}, EPIC-KITCHENS~\cite{damen2018scaling}, FineAction~\cite{liu2022fineaction}, Ego-Exo4D~\cite{grauman2024egoexo4d}, and MECCANO~\cite{ragusa2023meccano} provide complementary egocentric, fine-grained, and multimodal settings. We adopt ActivityNet because its action instances and ActivityNet Captions descriptions share the same untrimmed videos, providing a common temporal source for simulated-event recognition, auxiliary language supervision, and causal localization.

%-------------------------------------------------
\subsubsection{Event--Language Learning.}
Large-scale vision--language pretraining, exemplified by CLIP~\cite{radford2021learning}, established text supervision as a source of semantic structure. EventGPT~\cite{liu2025eventgpt} introduced an event-based MLLM using synthetic event--text pretraining and real-event instruction tuning. EventBench~\cite{liu2025eventbench} expanded evaluation across event understanding, recognition, and spatial reasoning, while EventFlash~\cite{liu2026eventflash} introduced the EventMind instruction corpus and efficient long-event processing. These works primarily target instruction learning, open-ended understanding, or MLLM evaluation. In contrast, Event~ActivityNet inherits action annotations and human-written descriptions from ActivityNet, derives temporally matched caption--action pairs, and uses text only as auxiliary supervision for annotated-segment recognition and causal online TAL. Further discussion appears in~\Cref{secA:rw_ext}.

%======================================================
\section{Event ActivityNet: Data and Protocols}
\label{sec:benchmark}

We present the construction and evaluation setup of \textbf{Event ActivityNet}, a large-scale simulated-event testbed for action recognition and online temporal action localization. The public release is derived from non-interpolated ActivityNet videos retaining per-video native nominal or average frame rates and is accompanied by fixed annotations, manifests, timing metadata, and integrity records.

\subsection{Benchmark Tasks and Supervision}
\label{sec:setup}
Given an untrimmed RGB video $V=\{I_t\}_{t=1}^{T}$ from ActivityNet Captions v1.3~\cite{KrishnaHRFN17}, whose videos come from ActivityNet v1.3~\cite{caba2015activitynet}, we construct a polarity-aware event-voxel sequence $X=\{X_q\}_{q=1}^{Q}$. Each tensor $X_q\in\mathbb{R}^{B\times H\times W}$ contains $B$ temporal bins. We consider two downstream tasks.

\subsubsection{Annotated-segment Action Recognition.}
For recognition, we select the target ActivityNet v1.3 action instance, extract its annotated temporal segment, and use the original v1.3 class as the recognition label. Thus, the model does not assign a single longest-instance label to an entire untrimmed video. This protocol isolates semantic event representation, while the complete stream is reserved for online TAL.

\subsubsection{Causal Online Temporal Action Localization.}
Online TAL predicts $K$ action instances as $\{(\hat{s}_k,\hat{e}_k,\hat{c}_k)\}_{k=1}^{K}$. At time $t$, all features, fusion operations, and predictions are computed from the observed prefix only, with no access to future observations and no post-hoc modification of past predictions. The original ActivityNet v1.3 instance-level annotations are used unchanged as localization ground truth.

\subsubsection{Caption--Action Temporal Alignment.}
ActivityNet Captions~\cite{KrishnaHRFN17} and ActivityNet v1.3~\cite{caba2015activitynet} use different temporal segmentations and annotation goals. For a caption segment $C_j=[s^c_j,e^c_j]$ and action annotations $\mathcal{A}=\{A_k=([s_k,e_k],y_k)\}_{k=1}^{K}$, we compute
\begin{equation}
\mathrm{tIoU}(C_j,A_k)=\frac{|[s^c_j,e^c_j]\cap[s_k,e_k]|}{|[s^c_j,e^c_j]\cup[s_k,e_k]|},
\end{equation}
and assign the caption to $k^\star=\arg\max_k\mathrm{tIoU}(C_j,A_k)$. Captions with maximum overlap below $\eta=0.1$ are marked unmatched and excluded from supervised alignment. The resulting matches provide only weak event--language supervision and do not alter recognition labels or TAL boundaries.

\subsection{Original-Rate Event-Voxel Construction}
\label{sec:pipeline}
Figure~\ref{fig:pipeline} summarizes the construction pipeline:
(i) source selection and annotation alignment,
(ii) direct original-rate video-to-voxel generation,
(iii) reconstruction-based perceptual quality control, and
(iv) packaging and validation.

\subsubsection{Dataset Curation.}
We curate the dataset from the 14{,}926-video ActivityNet Captions v1.3 train/validation pool~\cite{KrishnaHRFN17}, whose videos originate from ActivityNet v1.3~\cite{caba2015activitynet}. Curation follows three principles: coverage of all 200 primary action classes, duration stratification across short, middle-duration, and long videos, and motion/illumination-oriented enrichment based on caption triggers and first-frame brightness. The finalized subset contains 3{,}263 videos and 106.94 hours, including 2{,}316 training and 947 validation videos. The source videos are decoded without conversion to a common frame rate and retain different per-video nominal or average rates. Voxel construction follows decoded frame order; the released rational-rate and duration metadata provide an approximate frame-index-to-seconds mapping rather than recovering within-video presentation timestamps. Detailed curation statistics and source-to-subset distribution comparisons are provided in~\Cref{secA:subset_stats}.

\subsubsection{Rationale for Original-Rate Discrete Voxels.}
High-frame-rate interpolation attempts to infer temporal changes that were not captured by the source video and may introduce interpolation artifacts. V2V argues that discrete voxelization can discard unreliable intra-bin timing while preserving inter-bin dynamics derived from frame differences, and reports comparable downstream reconstruction results for interpolated and discrete voxel inputs~\cite{abs-2505-16797}. V2CE likewise distinguishes accumulation-based pipelines, where fine event timestamps are collapsed, from timestamp-sensitive architectures such as GNNs and SNNs that require continuous events~\cite{ZhangCCYDMR24}. Because Event ActivityNet targets voxel-based recognition and temporal localization, we generate discrete 5-bin and 9-bin tensors directly from adjacent decoded frames. This avoids inventing intermediate frames or storing continuous raw events, but it neither preserves within-video presentation timestamps nor recovers native microsecond evidence absent from the RGB source. Exact generator parameters and pseudocode are provided in~\Cref{secA:v2v_generation}, and a matched generator comparison is reported in~\Cref{secA:sim_compare}.

\subsubsection{Direct Video-to-Voxel Conversion.}
The finalized HDF5 payload is produced by a customized GPU implementation of
V2V~\cite{abs-2505-16797}. FFmpeg decodes each selected video directly to
8-bit grayscale frames at source resolution; the pipeline performs no resize,
crop, frame-rate conversion, intermediate 240-Hz interpolation, or creation of
a timestamped event list. Adjacent log-intensity changes are accumulated with
fixed contrast thresholds $c^+=c^-=0.2$ into dense signed transition slices
$q_i$. For $B\in\{5,9\}$, one adjacent-frame transition is assigned to each
voxel bin,
\begin{equation}
X_{t,b}=q_{Bt+b}.
\label{eq:main_voxel_grouping}
\end{equation}
A full tensor therefore spans $B$ transitions and $B+1$ source frames; an
incomplete final tensor is zero-padded and stored as int16. These tensors are
discrete simulated-event representations rather than native asynchronous
event lists. The complete state update, noise and hot-pixel settings, temporal
indexing convention, parameter table, and pseudocode are provided in
\Cref{secA:v2v_generation}; the matched generator comparison is reported in
\Cref{secA:sim_compare}.

\subsubsection{Reconstruction-Based Perceptual Quality Audit.}
For every annotated action instance, we center a voxel window at the segment
midpoint, reconstruct it with the fixed pretrained HyperE2VID
model~\cite{ercan2024hypere2vid}, and compare one center reconstruction with
the nominally aligned source frame using LPIPS~\cite{zhang2018unreasonable}.
The video-level score is the maximum over its annotated action centers. We use
$\delta=0.30$ as a soft audit threshold rather than an exclusion rule, so all
3{,}263 structurally validated videos remain in the benchmark. This diagnostic
measures retained reconstructable content at action centers only; it does not
fully validate temporal boundaries, background-to-action transitions, native
asynchronous statistics, sensor noise, or microsecond timing. Exact frame
mapping, reconstructor and LPIPS settings, aggregation, and threshold
sensitivity are detailed in \Cref{secA:lpips_sensitivity}, while
post-generation voxel statistics appear in \Cref{secA:event_diagnostics} of
the appendix.

\subsection{Dataset Scales and Release Structure}
\label{sec:release}
Event ActivityNet defines nested development scales selected from the same
Large collection: a fixed approximately 10-hour diagnostic subset, Small
(approximately 20 hours), Medium (approximately 50 hours), and Large
(3{,}263 videos, 106.94 hours). These scales reuse the same payload through
ID manifests rather than duplicating data, and the 10-hour subset is not a
separate raw-event release.

The benchmark provides matched 5-bin and 9-bin HDF5 representations with
identical video IDs, annotations, timing records, and split definitions,
together with fixed manifests, annotation-issue records, structural
validation, and shard-level checksums. The 9-bin payload follows the same
release organization and is evaluated with the 5-bin representation in
\Cref{sec:scale_bin_transfer}. Detailed HDF5 fields, archive sizes, shard
counts, checksum inventory, metadata files, and integrity checks appear in
\Cref{secA:release_integrity}.

%======================================================
\section{Reference Models}
\label{sec:baselines}

We establish reference models for \emph{untrimmed} event streams that address three challenges: (i) nonstationary event dynamics that make fixed-window slicing brittle, (ii) semantic grounding through captions and class prompts, and (iii) controlled comparison of RGB, event, and fused RGB--event cues under causal online TAL.

\subsubsection{Scope of the Reference Architectures.}
The benchmark and its protocols are our primary contribution. We therefore adapt ExACT~\cite{ZhouZLW24}, SimOn~\cite{abs-2211-04905}, and MATR~\cite{song2024online}, rather than introduce new backbone families. For recognition, ExACT is augmented with voxel-native adaptive framing and caption alignment. For online TAL, SimOn is evaluated with matched RGB-only, event-only, and causally fused RGB--event inputs; the event and fusion variants are also integrated with MATR's memory-based instance decoder and optional boundary refinement. Architecture diagrams and adaptation details appear in~\Cref{secA:baseline_architectures}.

%------------------------------------------------------
\subsection{Adaptive Event Framing via Split--Merge}
\label{sec:AEF}

Fixed windows are brittle for untrimmed event streams containing bursts, silent intervals, and abrupt motion or illumination changes. Our \textbf{AEF-Split\&Merge (AEF-SM)} is a lightweight nonparametric procedure that represents each window $W$ by its voxel activity rate $\rho(W)$ and normalized polarity-aware spatial histogram $\mathbf{h}(W)$. Their dissimilarity combines a log activity-rate ratio with Jensen--Shannon divergence:
\begin{equation}
\begin{aligned}
D(W_a,W_b)
={}&
\omega_\rho
\left|
\log
\frac{\rho(W_a)+\epsilon}
     {\rho(W_b)+\epsilon}
\right|\\
&+
\omega_h
D_{\mathrm{JS}}\!\left(
\mathbf{h}(W_a),\mathbf{h}(W_b)
\right).
\end{aligned}
\label{eq:main_aef_distance}
\end{equation}
AEF-SM splits a window at the pivot of maximum internal dissimilarity when the score exceeds $\tau_{\mathrm{split}}$ and both resulting windows satisfy the minimum-duration constraint. It then greedily merges adjacent windows with dissimilarity below $\tau_{\mathrm{merge}}$. Statistics, pivot selection, threshold calibration, default parameters, and the full algorithm are detailed in~\Cref{secA:event_ar,secA:algor}.

%------------------------------------------------------
\subsection{Event--Text Contrastive Alignment}
\label{sec:triple_align}

Event~ActivityNet provides class prompts and sample-level captions. Following ExACT~\cite{ZhouZLW24}, we train an event encoder against a frozen text encoder using handcrafted and learnable class prompts and temporally matched ActivityNet captions. For sample $i$, let $P_i$ be the similarity mass of its positives: the class prototype and, when available, an eligible matched caption. The normalization $Z_i$ also includes in-batch class prototypes and downweighted eligible caption negatives. For a batch of size $n_{\mathrm{b}}$,
\begin{equation}
\mathcal{L}_{\mathrm{con}}
=
-\frac{1}{n_{\mathrm{b}}}
\sum_{i=1}^{n_{\mathrm{b}}}
\log\frac{P_i}{Z_i}.
\label{eq:lalign_ecp}
\end{equation}
Unmatched captions are excluded, and caption supervision does not alter inherited labels or temporal boundaries. We compare prompt-only, caption-only, and combined objectives for $\eta\in\{0.1,0.3,0.5\}$, as well as a positive-only Smooth-$\ell_1$ alternative inspired by RECON~\cite{qi2023recon}. Objective definitions and caption ablations appear in~\Cref{secA:event_text_objective,secA:caption_ablation}.

%------------------------------------------------------
\subsection{Causal RGB--Event Fusion for Online TAL}
\label{sec:ontal_fusion}

We provide SimOn-based~\cite{abs-2211-04905} online TAL baselines whose streaming heads use only the observed prefix. The TAL head is fixed across matched RGB-only, event-only, and RGB--event configurations to isolate modality contributions. C1 downsamples event features to the RGB rate, C2 replicates RGB features at the event rate, and cross-attention (CA) uses event-prefix queries with RGB-prefix keys and values:
\begin{equation}
\mathbf{F}_{\le t}
=
\operatorname{CA}\!\left(
\mathbf{E}_{\le t},
\mathbf{R}_{\le t},
\mathbf{R}_{\le t};
\mathbf{M}_{\mathrm{causal}}
\right).
\label{eq:main_causal_fusion}
\end{equation}
The matched RGB-only control retains the CA model's RGB encoder, timestamp alignment, projection dimension, training schedule, and TAL head, but excludes event inputs. All fusion operations and caches are monotonic and prefix-only.

We also adapt MATR~\cite{song2024online}, retaining its selective memory queue and separate start/end decoders with the same event-only or causally fused features. The strongest variant adds lightweight residual corrections to coarse boundary predictions. Temporal alignment, tensor shapes, resampling, causal caching, architectures, and boundary refinement are detailed in~\Cref{secA:event_ontal,secA:arch_simon,secA:arch_matr}.

%=================================================
\section{Evaluation of Reference Models}
\label{sec:exp}

We evaluate reference models on two Event~ActivityNet tasks: event-based action recognition and causal online temporal action localization (Online TAL). We aim to provide strong, reproducible reference points rather than exhaustively tune each model. All reported results are 3-run means unless noted otherwise. All configurations use fixed training schedules and report the final checkpoint; the 947-video validation split is reserved for final evaluation, not early stopping or checkpoint selection. Provenance IDs for the core recognition, alignment, and Small-scale SimOn configurations appear in~\Cref{tab:protocol_summary,tab:protocol_ontal}; progressive-scale, native-transfer, and MATR protocols are specified with their respective experiments.

\subsection{Event-Based Action Recognition}
\label{sec:mar}

\noindent\textbf{Training Protocol.}
Unless stated otherwise, we follow ExACT's backbone and training
recipe~\cite{ZhouZLW24}. On Event~ActivityNet (Small), models are trained for
100 epochs with a fixed optimizer and learning-rate schedule. We report the
mean over three runs at the final scheduled checkpoint.

\noindent\textbf{Datasets and Metrics.}
The main paper reports Top-1 and Top-5 accuracy on Event~ActivityNet (Small,
5-bin). Results on PAF~\cite{MiaoCNZRBK19},
HARDVS~\cite{wang2024hardvs}, DVS128 Gesture~\cite{amir2017low}, and
SeAct~\cite{ZhouZLW24}, including AEF and alignment controls on public data,
appear in~\Cref{secA:public_recognition}.

\noindent\textbf{Voxel Inputs and Recognition Controls.}
To interface the released voxels with standard backbones, we accumulate
consecutive tensors until their voxel-event mass
$\sum_q\lVert X_q\rVert_1$ reaches 100k, then form
$\mathbf{x}=[\mathbf{x}^{+},\mathbf{x}^{-},
\mathbf{x}^{+}-\mathbf{x}^{-}]$. This threshold denotes cumulative $\ell_1$
mass in signed voxels, not events from a raw timestamped stream. We optionally
apply a $1{\times}1$ projection before the ExACT backbone. Because ExACT's AFE
module assumes raw events, we disable it and evaluate voxel-native AEF-SM with
the same backbone and schedule. We also compare positive-only Smooth-$\ell_1$
alignment with the full prompt--caption contrastive objective; prompt-only,
caption-only, combined, and alignment-threshold controls are reported
in~\Cref{tab:caption_supervision_ablation}.

\begin{table}[t]
\centering
\caption{\textbf{Action-recognition ablations on Event~ActivityNet (Small).}
Mean $\pm$ sample standard deviation over three runs using 5-bin voxels
and a fixed 100-epoch schedule. The final three rows include the
$1{\times}1$ projection; both alignment variants also use AEF-SM.}
\label{tab:rec_main}
\small
\renewcommand{\arraystretch}{1.05}
\resizebox{\columnwidth}{!}{
\begin{tabular}{lcc}
\toprule
\textbf{Setting} & \textbf{Top-1} (\%) & \textbf{Top-5} (\%) \\
\midrule
Base, w/o AFE & \meanstd{51.11}{0.15} & \meanstd{51.67}{0.19} \\
$+\,1{\times}1$ projection & \meanstd{52.56}{0.25} & \meanstd{56.24}{0.25} \\
$+$ AEF-SM & \meanstd{53.31}{0.21} & \meanstd{57.93}{0.24} \\
$+$ Positive-only Smooth-$\ell_1$ & \meanstd{51.22}{0.22} & \meanstd{55.23}{0.23} \\
$+$ Contrastive prompts + captions ($\eta=0.1$)
& \meanstdbf{57.24}{0.21}
& \meanstdbf{61.35}{0.23} \\
\bottomrule
\end{tabular}}
\end{table}

\subsubsection{Recognition Results.}
Event~ActivityNet is more challenging than short, actor-centered recognition
datasets: its annotated segments span 200 categories in unconstrained videos
with background and camera motion, compression, and nonstationary event
activity. The $1{\times}1$ projection raises Top-1/Top-5 accuracy from
51.11/51.67 to 52.56/56.24, and AEF-SM further improves it to 53.31/57.93.
The full prompt--caption objective achieves 57.24/61.35, whereas positive-only
alignment performs worse. Separate prompt, caption, combined, and threshold
controls are used to isolate the contribution of caption supervision.

\subsection{Causal Online Temporal Action Localization}
\label{sec:ontal}

\subsubsection{Task and Evaluation Protocol.}
Online TAL predicts action instances using only the observed prefix, thereby
evaluating early prediction under streaming constraints. We follow the setting
introduced by CAG-QIL~\cite{KangKKK21} and use SimOn~\cite{abs-2211-04905} as
our primary framework. Following the SimOn and MATR convention on
THUMOS14~\cite{abs-2211-04905,song2024online}, we report class-averaged mAP at
tIoU thresholds $\{0.3,0.4,0.5,0.6,0.7\}$. \emph{Avg. mAP} is the mean across
these thresholds, and \emph{mAP@30} denotes mAP at tIoU $0.3$. The published
RGB SimOn result on ActivityNet v1.3 uses different ground-truth construction
and thresholds $\{0.5,0.75,0.9\}$; we therefore cite it only as context and
exclude it from the modality comparison.

\subsubsection{Backbones and Event Encoders.}
We keep the SimOn localization head fixed while varying its input features.
Event-only models use ExACT~\cite{ZhouZLW24} or Group Event Transformer
(GET)~\cite{PengZX0023}; the matched RGB-only control reuses the multimodal
model's RGB encoder and temporally aligned RGB branch, followed by the same
input projection and SimOn head. We evaluate frozen encoders and two
fine-tuning settings: \texttt{w/ T} uses an encoder learning-rate multiplier of
0.1 relative to the TAL head, whereas the stronger \texttt{w/ TM} uses 0.5.
Simplified SimOn and MATR diagrams appear in
Figures~\ref{fig:arch_simon} and~\ref{fig:arch_matr}; complete Small- and
Medium-scale MATR results are reported in~\Cref{tab:matr_scales}.

\subsubsection{Streaming Step Size and Voxelization.}
Unless noted, we use 5-bin voxels and express streaming steps in seconds. Because videos retain heterogeneous nominal or average source rates, each voxel index spans a video-dependent duration. We map each step $\Delta\in\{0.25,0.5,1.0\}$ seconds to transition indices using the released rational-rate metadata rather than assuming a 240-Hz grid. Because the ActivityNet boundaries remain in seconds while the released voxels are indexed by decoded-frame order, this conversion is exact in frame-index units but approximate for within-video variable-frame-rate streams. \Cref{tab:ontal_main} compares step sizes, event encoders, tuning settings, and RGB--event fusion.

\subsubsection{Matched Modality Controls and RGB--Event Fusion.}
The RGB-only control uses the multimodal model's source-RGB branch, aligns its
features with the event-query timestamps, projects them to the same head
dimension, and follows the same schedule and final-checkpoint rule as the
marked event-only and cross-attention models. We evaluate concatenation using
\texttt{C1}, which downsamples events to the RGB rate, and \texttt{C2}, which
replicates RGB features by $N$ to match the higher event rate; for example,
\texttt{(1s, 10)} yields 0.1-s-equivalent alignment. Cross-attention
(\texttt{CA}) instead uses events as queries and RGB features as keys and
values. The marked RGB-only, event-only, and CA rows in~\Cref{tab:ontal_main}
form the matched modality comparison.

\begin{table}[t]
\centering
\caption{\textbf{Causal Online TAL on Event~ActivityNet (Small).}
All models use 5-bin voxels. Avg. mAP averages tIoU thresholds
$\{0.3,0.4,0.5,0.6,0.7\}$; mAP@30 denotes tIoU $0.3$.
Rows marked $^{\ddagger}$ form the matched modality comparison and report
mean $\pm$ sample standard deviation over three runs; other rows are
3-run means.}
\label{tab:ontal_main}
\small
\renewcommand{\arraystretch}{1.05}
\resizebox{\columnwidth}{!}{
\begin{tabular}{lccc}
\toprule
\textbf{Configuration} & \textbf{Modality} & \textbf{Avg.\ mAP} & \textbf{mAP@30} \\
\midrule
\multicolumn{4}{c}{Event Only: Streaming Step}\\
\midrule
SimOn (step $=0.25$\,s) & Event & 12.4 & 23.1 \\
SimOn (step $=0.5$\,s) & Event & 11.3 & 21.6 \\
SimOn (step $=1.0$\,s) & Event & 15.3 & 25.5 \\
\midrule
\multicolumn{4}{c}{Event Only: Encoder and Fine-Tuning}\\
\midrule
SimOn + ExACT encoder~\cite{ZhouZLW24} & Event & 17.7 & 28.9 \\
SimOn + GET encoder~\cite{PengZX0023} & Event & 18.3 & 32.7 \\
SimOn + GET (w/ T)$^{\ddagger}$ & Event & \meanstd{20.8}{0.6} & \meanstd{31.5}{0.9} \\
SimOn + GET (w/ TM) & Event & \textbf{22.5} & \textbf{35.3} \\
\midrule
\multicolumn{4}{c}{Matched RGB-Only Control}\\
\midrule
SimOn + RGB encoder$^{\ddagger}$ & RGB & \meanstd{24.2}{0.6} & \meanstd{38.3}{1.1} \\
\midrule
\multicolumn{4}{c}{RGB--Event Fusion}\\
\midrule
SimOn + GET (w/ T) + C1 & RGB--Event & 22.5 & 35.3 \\
SimOn + GET (w/ T) + C2 (1s, 10) & RGB--Event & 24.3 & 39.9 \\
SimOn + GET (w/ T) + C2 (0.5s, 5) & RGB--Event & 23.6 & 37.4 \\
SimOn + GET (w/ T) + CA$^{\ddagger}$ & RGB--Event
& \meanstdbf{25.8}{0.7}
& \meanstdbf{40.6}{1.0} \\
\bottomrule
\end{tabular}
}

{\scriptsize\raggedright
$^{\ddagger}$Matched rows share the split, streaming timestamps, SimOn head,
training budget, final-checkpoint rule, and evaluator; their variation is the
sample standard deviation across three runs. The RGB-only model reuses the CA
model's RGB branch without event input.\par}
\end{table}

\subsubsection{Online TAL Results.}
The nonmonotonic step-size trend reflects a trade-off between boundary
resolution and feature stability: shorter steps preserve temporal precision
but may contain sparse or bursty evidence, whereas longer steps aggregate more
stable motion statistics. Stronger encoders and fine-tuning improve event-only
performance. Under the matched protocol, causal cross-attention reaches
25.8 Avg. mAP and 40.6 mAP@30, exceeding RGB only by 1.6/2.3 points and the
matched event-only model by 5.0/9.1 points. RGB appearance thus provides
substantial context, while event cues offer an additional measurable gain.

%\subsection{Scale, Voxel Granularity, and Native-Event Transfer}
\subsection{Scale, Granularity, and Transfer}
\label{sec:scale_bin_transfer}

\subsubsection{Progressive Scaling.}
We evaluate fixed nested subsets with identical model configurations. The
diagnostic, Small, Medium, and Large scales contain 342, 667, 1{,}537, and
3{,}263 videos (approximately 10, 20.00, 50.00, and 106.94 hours). All scales
use 100 recognition epochs and 16 Online TAL epochs. Diagnostic and Small
start from random initialization, whereas Medium
and Large independently start from the same Small checkpoint. Recognition
rises from 52.25/57.35 to 66.42/73.41 in Top-1/Top-5, while online TAL improves
from 21.7/32.9 to 29.0/46.3 in Avg. mAP/mAP@30. Because larger scales reuse
the Small checkpoint and fixed epochs entail more optimization steps, this is
a practical scaling study rather than a controlled scaling law; full
memberships and training lineage appear in~\Cref{secA:scale_protocol}.

\begin{figure}[t]
    \centering
    \includegraphics[width=\columnwidth]{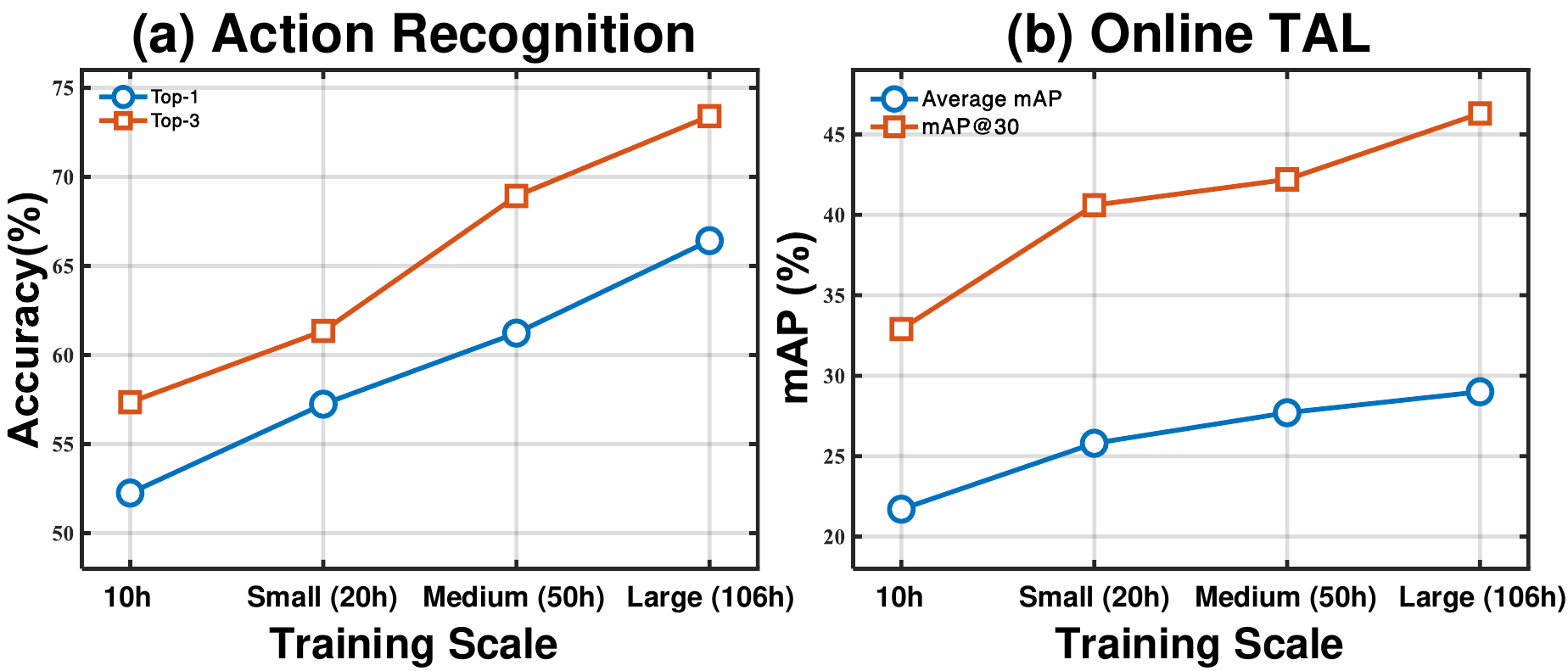}
    \caption{\textbf{Performance across Event~ActivityNet scales.}
    Recognition and causal online TAL on fixed nested subsets of 342, 667,
    1{,}537, and 3{,}263 videos. Diagnostic and Small use random
    initialization; Medium and Large independently start from the Small
    checkpoint.}
    \label{fig:scale_curve}
\end{figure}

\subsubsection{Effect of Voxel Granularity.}
The 5-bin and 9-bin settings share source videos, annotations, splits,
generator, backbones, and optimization, but group different numbers of frame
transitions per tensor and therefore have different sequence lengths and
temporal spans. They are matched in data membership and training settings, not
index-wise time. Recognition changes by at most 0.21 points. For Online TAL,
the alternative 9-bin grouping gains 0.2--0.3 Avg. mAP and 0.6--1.8 mAP@30,
with the largest change for causal RGB--event fusion. Thus, voxel grouping can
affect localization metrics even when recognition is stable. This comparison
tests grouping robustness and does not recover native temporal evidence;
additional results appear in~\Cref{secA:bin_compare}.

\begin{table}[t]
\centering
\caption{\textbf{Effect of voxel granularity on Event~ActivityNet (Small).}
All settings use identical videos, annotations, splits, and training settings;
only the voxelization granularity changes. Recognition reports Top-1/Top-5
accuracy; online TAL reports Avg. mAP/mAP@30.}
\label{tab:bin_compare_main}
\small
\resizebox{\columnwidth}{!}{
\begin{tabular}{llcc}
\toprule
\textbf{Task} & \textbf{Setting} & \textbf{5-bin} & \textbf{9-bin} \\
\midrule
Recognition & $1{\times}1$ projection & 52.56 / 56.24 & 52.57 / 56.39 \\
Recognition & AEF-SM + contrastive & 57.24 / 61.35 & 57.32 / 61.56 \\
Online TAL & GET (w/ TM) & 22.5 / 35.3 & 22.7 / 35.9 \\
Online TAL & GET (w/ T) + CA & 25.8 / 40.6 & 26.1 / 42.4 \\
\bottomrule
\end{tabular}}
\end{table}

\subsubsection{Transfer to Native Event Recognition.}
We compare target-only scratch (native data only), joint scratch (simulated and
native data trained jointly), and staged EA pretraining followed by native-data
fine-tuning under matched native-data budgets. Joint scratch and EA pretraining
use the same simulated and native samples but differ in joint versus staged
training, separating extra-data exposure from the effect of training order.
\begin{table}[t]
\centering
\caption{\textbf{Controlled transfer to native event-recognition datasets.}
Top-1/Top-5 accuracy using 10\%, 50\%, or 100\% of each target dataset.
\emph{Target scratch} uses only native data, \emph{Joint scratch} jointly uses
Event~ActivityNet and native data, and \emph{EA pretrain} fine-tunes an
Event~ActivityNet-pretrained model. The transferred event backbone receives no external visual pretraining.}
\label{tab:transfer_controlled}
\small
\resizebox{\columnwidth}{!}{
\begin{tabular}{lcccc}
\toprule
\textbf{Target} & \textbf{Training} &
\textbf{10\%} & \textbf{50\%} & \textbf{100\%} \\
\midrule
\multirow{3}{*}{HARDVS}
& Target scratch
& 84.83 / 92.81
& 88.25 / 95.24
& 89.74 / 96.36 \\
& Joint scratch
& 86.70 / 94.30
& 89.45 / 96.20
& 90.63 / 97.02 \\
& EA pretrain
& \textbf{90.32 / 96.83}
& \textbf{91.53 / 98.02}
& \textbf{92.35 / 98.25} \\
\midrule
\multirow{3}{*}{SeAct (ExACT-category)}
& Target scratch
& 58.71 / 63.33
& 62.83 / 67.57
& 65.58 / 69.74 \\
& Joint scratch
& 60.80 / 66.20
& 64.90 / 69.40
& 66.67 / 70.78 \\
& EA pretrain
& \textbf{65.56 / 70.01}
& \textbf{68.53 / 75.46}
& \textbf{70.02 / 76.55} \\
\midrule
\multirow{3}{*}{SeAct (ExACT-caption)}
& Target scratch
& 58.71 / 66.82
& 64.06 / 71.92
& 66.34 / 74.35 \\
& Joint scratch
& 59.53 / 69.40
& 65.80 / 73.60
& 67.39 / 75.41 \\
& EA pretrain
& \textbf{66.36 / 73.14}
& \textbf{69.23 / 76.96}
& \textbf{69.23 / 77.73} \\
\bottomrule
\end{tabular}}
\end{table}
EA pretraining outperforms both scratch protocols for every dataset and
target-data fraction. Top-1 gains over target scratch range from 2.61 to 7.65
points and are generally largest at 10\% supervision; the advantage over joint
scratch suggests that staged pretraining is more effective than joint mixing
in these experiments.

\subsubsection{Localization Heads and Boundary Refinement.}
\Cref{tab:controlled_modality} compares SimOn and MATR on Medium using the same
causal cross-attended features, so the primary change is the localization head.
MATR introduces a selective memory queue and separate start/end decoding paths;
the offset variant additionally applies residual boundary correction.
\begin{table}[t]
\centering
\caption{\textbf{Localization-head robustness and boundary refinement on
Event~ActivityNet (Medium).}
SimOn and MATR receive the same causal cross-attended features. Entries
are mean $\pm$ sample standard deviation over three runs; the offset variant
further refines the coarse time predictions.}
\label{tab:controlled_modality}
\small
\resizebox{\columnwidth}{!}{
\begin{tabular}{llcc}
\toprule
\textbf{Head} & \textbf{Setting} & \textbf{Avg. mAP} & \textbf{mAP@30} \\
\midrule
SimOn & GET (w/ T) + CA & \meanstd{27.7}{0.3} & \meanstd{44.2}{0.7} \\
MATR & GET (w/ T) + CA & \meanstd{29.4}{0.4} & \meanstd{43.3}{1.0} \\
MATR & GET (w/ T) + CA + offset & \meanstdbf{29.9}{0.4} & \meanstdbf{46.6}{1.1} \\
\bottomrule
\end{tabular}}
\end{table}
MATR raises Avg. mAP from 27.7 to 29.4 but reduces mAP@30 from 44.2 to 43.3;
offset refinement improves both metrics to 29.9/46.6. Complete C1, C2, CA, and
offset results on Small and Medium appear in~\Cref{tab:matr_scales}.

%=================================================
\section{Conclusion}
\label{sec:conclusion}

We introduced \textbf{Event ActivityNet}, a scalable simulated-event testbed
for untrimmed action understanding. Reusing ActivityNet's human temporal
annotations and timestamped captions, it supports annotated-segment
recognition, auxiliary event--language supervision, and causal online TAL
without LLM-generated supervision. The release comprises matched 5- and
9-bin voxels for 3{,}263 videos (200 classes, 106.94 hours), plus fixed splits,
timing, annotation and alignment metadata, manifests, integrity records, and
shard-level checksums. Reference baselines cover adaptive framing,
prompt/caption alignment, matched RGB-only, event-only, and RGB--event Online
TAL, progressive scaling, voxel-granularity analysis, and limited native-event
validation. As a simulated-event testbed, it does not replace native-camera
data needed for deployment claims; see~\Cref{secA:limit}.

% Appendix for the public arXiv preprint.
\clearpage
\appendix

\renewcommand{\theequation}{S\arabic{equation}}
\setcounter{equation}{0}
\renewcommand{\thefigure}{S\arabic{figure}}
\setcounter{figure}{0}
\renewcommand{\thetable}{S\arabic{table}}
\setcounter{table}{0}

\section*{Table of Appendix Contents}

\appitem{Extended Related Work}{secA:rw_ext}
\appitem{Additional Component Analyses}{secA:add_ana}
\appsubitem{Implementation and Training Settings}{secA:implementation}
\appsubitem{Progressive-Scale Training Protocol}{secA:scale_protocol}
\appsubitem{Temporal Discretization: 5-Bin vs.\ 9-Bin}{secA:bin_compare}
\appsubitem{Direct Video-to-Voxel Generation}{secA:v2v_generation}
\appsubitem{Held-Out Generator Sensitivity}{secA:sim_compare}
\appsubitem{Caption--Action Alignment Quality}{secA:caption_quality}
\appsubitem{Controlled Caption-Supervision Ablation}{secA:caption_ablation}
\appsubitem{Event--Text Objective Details}{secA:event_text_objective}
\appsubitem{LPIPS Configuration and Sensitivity}{secA:lpips_sensitivity}
\appsubitem{Post-Generation Event Diagnostics}{secA:event_diagnostics}
\appsubitem{Subset Curation and Distribution Shift}{secA:subset_stats}
\appsubitem{Release Packaging and Integrity}{secA:release_integrity}
\appsubitem{Event-Based Action Recognition}{secA:event_ar}
\appsubitem{Public Event-Recognition Evaluations}{secA:public_recognition}
\appsubitem{Event-Based Online TAL}{secA:event_ontal}
\appitem{Reference Baseline Architectures}{secA:baseline_architectures}
\appsubitem{ExACT-Based Recognition Adaptation}{secA:arch_exact}
\appsubitem{SimOn-Based Online TAL Adaptation}{secA:arch_simon}
\appsubitem{MATR-Based Online TAL Adaptation}{secA:arch_matr}
\appsubitem{Scale-Wise SimOn/MATR Results}{secA:matr_scales}
\appitem{AEF-Split{\&}Merge Procedure}{secA:algor}
\appitem{Limitations}{secA:limit}

\section{Extended Related Work}
\label{secA:rw_ext}

\subsubsection{Early Event Vision and Representations.} 
Event cameras, popularized by the Dynamic Vision Sensor (DVS) \cite{lichtsteiner2008128}, measure per-pixel brightness changes asynchronously with microsecond temporal resolution. As reviewed in \cite{gallego2022event}, this sensing modality has enabled substantial progress on optical flow \cite{zhu2019unsupervised}, video reconstruction \cite{rebecq2019events}, and depth estimation. Common representations (time surfaces, event frames, and voxel grids) convert sparse event streams into tensorized inputs for standard architectures.

\subsubsection{Event-Based Video Understanding.} 
Beyond standalone recognition, recent work explores RGB--event fusion for complementary motion cues and robustness~\cite{GehrigRGHS21}, as well as pretraining strategies adapted from large-scale video representation learning. However, most studies still use trimmed clips with predefined boundaries, leaving long-horizon stream parsing underexplored.

\subsubsection{Online TAL and Streaming Multimodal Trends.} 
In RGB videos, online TAL has been formalized by frameworks such as CAG-QIL~\cite{KangKKK21} and SimOn~\cite{abs-2211-04905}. More broadly, streaming video understanding is moving toward multimodal settings, including hierarchical vision--language streaming understanding~\cite{abs-2509-12145} and predictive future modeling for online audio-visual event parsing~\cite{YuPreFM25}. Event ActivityNet brings this evaluation style to causal protocols for untrimmed event streams.

\subsubsection{Synthetic Event Generation and Validation.}
Because collecting large-scale temporally annotated native event data is costly, video-derived simulation is increasingly used for representation learning and controlled evaluation. Event ActivityNet follows an original-rate direct video-to-voxel release pipeline rather than generating intermediate 240-Hz videos or storing continuous raw events. Event-to-video reconstruction methods, including HyperE2VID~\cite{ercan2024hypere2vid}, provide a practical post-generation diagnostic: we reconstruct the center frame of every annotated action segment and compare it with the corresponding source RGB frame using LPIPS~\cite{zhang2018unreasonable}. This evaluates retained reconstructable content but does not establish equivalence to native event-camera statistics. HDF5 structure, timing metadata, manifests, and checksums provide separate integrity checks.

\subsubsection{Event--Language Models and Benchmarks.} 
CLIP~\cite{radford2021learning} and related contrastive pretraining paradigms established strong visual--text alignment and prompt-based evaluation in frame-based vision. EventGPT~\cite{liu2025eventgpt} develops an event-based MLLM with synthetic event--text pretraining and real-world instruction tuning. EventBench~\cite{liu2025eventbench} provides a comprehensive MLLM benchmark spanning eight understanding, recognition, and spatial-reasoning tasks. EventFlash~\cite{liu2026eventflash} introduces EventMind, an instruction corpus with more than 500{,}000 samples, and studies efficient long-range event MLLM processing through token sparsification. Event ActivityNet is complementary: it inherits human temporal action boundaries and human-written captions from ActivityNet, derives temporal caption--action matches, and focuses on untrimmed streams and causal online localization rather than question answering or instruction following.

\section{Additional Component Analyses}
\label{secA:add_ana}

\subsubsection{Protocol Overview.} 
To improve reproducibility and make the core cross-task comparisons transparent, \Cref{tab:protocol_summary,tab:protocol_ontal} separate the recognition/alignment and Small-scale SimOn provenance IDs into two readable summaries. They record each setting's input, backbone or head, training recipe, and defining implementation detail. Progressive-scale training, native-event transfer, and MATR experiments use separate protocols described in their dedicated sections.

\subsubsection{Additional Sensitivity Analyses.} 
Beyond the main-paper results on the 5-bin voxel setting, we provide additional analyses to test whether the benchmark conclusions are sensitive to implementation choices in the event representation pipeline. In particular, we first compare representative best-performing settings under 5-bin and 9-bin voxelization to assess temporal discretization sensitivity. Since the goal here is to verify robustness rather than introduce new model variants, we keep the same split, backbone, and training protocol, and vary only the voxel discretization.

% -------------------- Protocol / provenance summaries --------------------
\begin{table*}[t]
\centering
\caption{\textbf{Recognition and event--text alignment protocol summary.}
Unless stated otherwise, Event ActivityNet rows use the Small split, 5-bin
voxels, final-epoch reporting, and 3-run means. R4 follows each public
benchmark's own ExACT protocol.}
\label{tab:protocol_summary}
\small
\setlength{\tabcolsep}{4pt}
\renewcommand{\arraystretch}{1.12}
\begin{tabular}{@{}p{0.05\textwidth}p{0.23\textwidth}p{0.15\textwidth}p{0.22\textwidth}p{0.27\textwidth}@{}}
\toprule
\textbf{ID} &
\textbf{Input / variant} &
\textbf{Backbone} &
\textbf{Training} &
\textbf{Defining detail} \\
\midrule
R1 &
Voxel$\rightarrow$frame (100k voxel-event mass/frame) &
ExACT &
100 epochs &
AFE disabled; no channel projection or AEF-SM \\
R2 &
R1 + $1{\times}1$ projection &
ExACT &
100 epochs &
AFE disabled; no AEF-SM \\
R3 &
R2 + AEF-SM &
ExACT &
100 epochs &
Voxel-native adaptive framing; parameters in Sec.~\ref{secA:event_ar} \\
R4 &
PAF / HARDVS / SeAct / DVS128 inputs &
ExACT &
Dataset-specific published schedule &
Independent public-dataset evaluation \\
\midrule
A1 &
R3 + positive-only Smooth-$\ell_1$ &
ExACT &
Same as R3 &
Frozen class/caption targets; no negatives \\
A2 &
R3 + unified contrastive alignment &
ExACT &
Same as R3 &
Class prompts + captions at $\eta=0.1$ \\
A3 &
R3 + class prompts only &
ExACT &
Same as R3 &
Caption terms removed \\
A4 &
R3 + captions only &
ExACT &
Same as R3 &
Class-prototype alignment removed; $\eta=0.1$ \\
A5 &
R3 + prompts and captions &
ExACT &
Same as R3 &
Threshold sweep $\eta\in\{0.1,0.3,0.5\}$ \\
\bottomrule
\end{tabular}
\end{table*}

\begin{table*}[t]
\centering
\caption{\textbf{Small-scale causal Online TAL protocol summary.}
All rows use the Event ActivityNet Small split, final-checkpoint reporting,
3-run means, and prefix-only inference.}
\label{tab:protocol_ontal}
\small
\setlength{\tabcolsep}{4pt}
\renewcommand{\arraystretch}{1.12}
\begin{tabular}{@{}p{0.05\textwidth}p{0.23\textwidth}p{0.18\textwidth}p{0.22\textwidth}p{0.23\textwidth}@{}}
\toprule
\textbf{ID} &
\textbf{Input / variant} &
\textbf{Encoder / head} &
\textbf{Training} &
\textbf{Defining detail} \\
\midrule
T1 &
Event-only; $\Delta\in\{0.25,0.5,1.0\}$ s &
SimOn &
16 epochs &
Streaming-step sweep with prefix-only inference \\
T2 &
Event-only; frozen or fine-tuned &
ExACT + SimOn &
16 epochs &
\texttt{w/ T}: LR$\times0.1$; \texttt{w/ TM}: LR$\times0.5$ \\
T3 &
Event-only; frozen or fine-tuned &
GET + SimOn &
16 epochs &
Same \texttt{w/ T}/\texttt{w/ TM} definitions \\
\midrule
F0 &
RGB-only, aligned to CA timestamps &
RGB encoder + SimOn &
16 epochs &
Same RGB branch/projection as F3; event input removed \\
F1 &
RGB+Event concatenation (C1) &
GET + SimOn &
16 epochs &
Events downsampled to the RGB rate \\
F2 &
RGB+Event concatenation (C2) &
GET + SimOn &
16 epochs &
RGB replicated by $N$; e.g., (1s,10), (0.5s,5) \\
F3 &
RGB+Event cross-attention (CA) &
GET + SimOn &
16 epochs &
Event queries; RGB keys/values; causal mask \\
\bottomrule
\end{tabular}
\end{table*}

\subsubsection{Checkpoint Reporting Rule.}
For all Event ActivityNet configurations, the training duration is fixed before evaluation and the final scheduled checkpoint is reported. Validation metrics are not used for early stopping or checkpoint selection. Within a controlled comparison, all variants use the same training duration and reporting rule.

\subsection{Implementation and Training Settings}
\label{secA:implementation}

\subsubsection{Environment.}
The dataset utilities and model experiments are implemented in Python~3 and
PyTorch~2.5.1 with CUDA~12.1. Experiments are run on a server equipped with
eight NVIDIA RTX~6000 Ada Generation GPUs (48\,GB memory per GPU), two
Intel Xeon Gold~5418Y processors (48 physical cores and 96 hardware threads in
total), and 503\,GiB of system memory. The server runs Ubuntu~22.04.4 LTS with
Linux kernel~5.15.0-185. The accompanying code/data package requires NumPy and
h5py to load and inspect the released HDF5 excerpts, while FFmpeg is used for
source-video decoding. HyperE2VID and LPIPS are used through their fixed
released implementations without benchmark-specific fine-tuning. All recognition and Online TAL results reported in this paper, including
the native-event transfer experiments, are averaged over three independent runs. Unless stated otherwise, we inherit the published
optimization and model settings of ExACT~\cite{ZhouZLW24},
SimOn~\cite{abs-2211-04905}, and MATR~\cite{song2024online}; our changes are
restricted to the voxel interface, event--text supervision, causal RGB--event
feature construction, and boundary refinement described in the paper.

\subsubsection{ExACT-Based Recognition.}
We use Adam with initial learning rate $10^{-5}$, weight decay
$2\times10^{-4}$, and cosine annealing to a minimum learning rate of
$10^{-6}$. The effective mini-batch size is 16. Event ActivityNet recognition
uses a fixed 100-epoch schedule; the public recognition evaluations retain the
dataset-specific ExACT schedule (100 epochs for PAF and SeAct and 25 epochs
for HARDVS). The frozen text encoder, final-checkpoint reporting rule, and
3-run mean for Event ActivityNet recognition are kept as specified in the
main paper. The original raw-event AFE front end is not used because our
release contains voxel tensors; it is replaced by AEF-SM where indicated.

\subsubsection{SimOn Head.}
The SimOn head uses four Transformer blocks, eight attention heads per block,
and dropout 0.1. We train for 16 epochs with effective batch size 256 using
AdamW and a step learning-rate scheduler with step size 3. The base learning
rates are $10^{-4}$ for the On-TAL Transformer and $10^{-5}$ for the context
embedding generator. For rows that fine-tune the event encoder, \texttt{w/ T}
and \texttt{w/ TM} use learning-rate multipliers 0.1 and 0.5 relative to the
TAL head, respectively. All modality variants retain the same SimOn head and
training schedule, and every reported Event ActivityNet Online TAL entry is
averaged over three runs.

\subsubsection{MATR Head.}
We use the THUMOS14-scale MATR configuration: segment length 64, feature
dimension 1024, three memory-encoder Transformer layers with eight attention
heads, flag threshold 0.5, memory size 7, five decoder layers with four
attention heads, and 10 paired class/boundary queries. Training uses Adam with
initial learning rate $10^{-8}$ and cosine annealing, effective batch size 64,
focal-loss parameters $\alpha=0.25$ and $\gamma=2$, and unit weights for all
classification, boundary, DIoU, and flag losses. Event-only and RGB--event
variants change only the causal input-feature construction; the optional
residual offset heads are described in \Cref{secA:event_ontal}.

\subsection{Progressive-Scale Training Protocol}
\label{secA:scale_protocol}

Each scale is defined by a fixed released ID manifest, and experiments use
these memberships directly without resampling. The diagnostic manifest is
nested within Small, Small within Medium, and Medium within Large. All scales
use the same model configurations and schedules; the only data-side change is
the manifest membership. Recognition uses 100 epochs at every scale, and
Online TAL uses 16 epochs at every scale. Diagnostic and Small start from
random initialization. Medium and Large are trained independently from the
same Small checkpoint rather than sequentially initializing Large from Medium.

\begin{table}[t]
\centering
\caption{\textbf{Fixed nested-scale training protocol.}
Medium and Large share the same Small-scale initialization but are trained
independently on their respective manifests.}
\label{tab:scale_protocol}
\small
\renewcommand{\arraystretch}{1.08}
\resizebox{\columnwidth}{!}{
\begin{tabular}{lccccc}
\toprule
\textbf{Scale} &
\textbf{Videos} &
\textbf{Hours} &
\textbf{Recognition} &
\textbf{Online TAL} &
\textbf{Initialization} \\
\midrule
Diagnostic & 342 & $\sim$10 & 100 epochs & 16 epochs & Random \\
Small & 667 & 20.00 & 100 epochs & 16 epochs & Random \\
Medium & 1{,}537 & 50.00 & 100 epochs & 16 epochs & Small checkpoint \\
Large & 3{,}263 & 106.94 & 100 epochs & 16 epochs & Small checkpoint \\
\bottomrule
\end{tabular}
}
\end{table}

\subsubsection{Note on Voxel Discretization.} 
All IDs in Table~\ref{tab:protocol_summary} refer to the 5-bin setting used in the main paper. For the appendix 9-bin sensitivity analysis, we reuse the same configurations and change only the voxelization granularity $B$.

\subsection{Temporal Discretization: 5-Bin vs.\ 9-Bin}
\label{secA:bin_compare}

Table~\ref{tab:bin_compare_main} in the main paper reports representative controlled comparisons. All runs reuse the same source videos, annotations, split, generator implementation, backbone, optimization, and random-seed policy; only the voxelization granularity $B$ changes. Here, ``matched'' does not mean index-wise temporal alignment: because each tensor groups $B$ consecutive frame transitions, 5-bin and 9-bin tensors have different sequence lengths and per-tensor temporal spans. The near-identical recognition results and stable method ranking indicate robustness to temporal discretization. The larger improvement in mAP@30 for cross-attention suggests that finer simulated bins can help boundary-sensitive fusion, although this should not be interpreted as recovery of native microsecond timing.

\subsection{Direct Video-to-Voxel Generation}
\label{secA:v2v_generation}

The finalized HDF5 tensors were produced by the customized path
\texttt{activitynet.py}
$\rightarrow$
\texttt{mp4\_to\_h5\_stream}
$\rightarrow$
\texttt{EventEmulatorGPU.video\_to\_voxel}. No learned checkpoint is loaded.
FFmpeg decodes the selected video stream to 8-bit grayscale frames at its
decoded spatial resolution; the generation path contains no resizing,
cropping, padding, spatial interpolation, or frame-rate conversion. The exact
RGB-to-gray coefficients are delegated to the FFmpeg build and source-stream
metadata.

For pixel $x$ in frame $G_i$, the intensity and log intensity are
\begin{equation}
\begin{aligned}
I_i(x)
&=
\operatorname{clamp}\!\left(
\frac{G_i(x)}{255},0,1
\right)^{2.2},\\
L_i(x)
&=
\log\!\left(0.001+I_i(x)\right).
\end{aligned}
\label{eq:supp_v2v_log}
\end{equation}
Within one simulator batch,
\begin{equation}
\begin{aligned}
\delta_i(x)
&=
L_{i+1}(x)-L_i(x)+\epsilon_i(x)+h(x),\\
s_i(x)
&=
p_0(x)+\sum_{j=0}^{i}\delta_j(x).
\end{aligned}
\label{eq:supp_v2v_state}
\end{equation}
where $\epsilon_i(x)\sim\mathcal{N}(0,0.1^2)$. A hot-pixel mask is sampled
independently per pixel with probability $0.001$; selected pixels receive
$h(x)\sim\mathcal{N}(0,0.1^2)$, held constant only within that simulator
batch. A device-specific PyTorch generator is reset to seed 42 on every
simulator call. Consequently, equal-shaped calls may reuse the same
pseudorandom sequence for the initial residual, base noise, and hot-pixel
sampling; the stochastic perturbations should therefore be interpreted as a
deterministic simulator artifact rather than independent sensor noise across
videos. Exact replay additionally requires the same device backend, software
stack, decoded frames, resolution, and decode-batch partition.

Let $c^+=c^-=0.2$ and
\begin{equation}
K_i^+=\left\lfloor\frac{\max(s_i,0)}{c^+}\right\rfloor,
\qquad
K_i^-=\left\lfloor\frac{\max(-s_i,0)}{c^-}\right\rfloor .
\label{eq:supp_v2v_levels}
\end{equation}
The signed slice for adjacent-frame transition $i$ is
\begin{equation}
q_i=(K_i^+-K_{i-1}^+)-(K_i^--K_{i-1}^-),
\label{eq:supp_v2v_slice}
\end{equation}
where $K_{-1}^{+/-}$ is computed from the incoming residual $p_0$. The first
batch initializes $p_0\sim\mathcal{U}[-0.2,0.2)$ per pixel. At the end of a
batch, the final state is reduced modulo the applicable positive or negative
threshold and carried into the next decode batch. Positive and negative
counts are not stored separately. No refractory period, leakage, shot-noise
process, timestamp jitter, or per-pixel threshold mismatch is included.

For $N$ decoded frames, the simulator yields $N-1$ transition slices. With
$B\in\{5,9\}$ and \texttt{frames\_per\_bin}$=1$,
\begin{equation}
\texttt{events}[t,b]=q_{Bt+b},
\qquad
T=\left\lceil\frac{N-1}{B}\right\rceil ,
\label{eq:supp_v2v_grouping}
\end{equation}
for available indices. Thus, a full $B$-bin voxel uses $B$ adjacent-frame
transitions and $B+1$ source frames. Consecutive voxels share a boundary frame
but no transition interval. A final incomplete voxel stores remaining slices
in its leading bins and zero-pads the trailing bins. Values are accumulated
in int32, clipped to $[-32768,32767]$, and stored as int16 without
normalization.

The generator indexes time by decoded frame order and does not consume
per-frame presentation timestamps. For rational average rate
$f=f_{\mathrm{num}}/f_{\mathrm{den}}$, voxel $t$ covers transition range
$[Bt,\min(Bt+B,N-1))$ and has approximate interval
\begin{equation}
\left[
\frac{Bt f_{\mathrm{den}}}{f_{\mathrm{num}}},
\frac{\min(Bt+B,N-1)f_{\mathrm{den}}}{f_{\mathrm{num}}}
\right].
\label{eq:supp_v2v_time}
\end{equation}
This mapping is exact in frame-index units but approximate in seconds for
within-video variable-frame-rate streams because per-frame presentation
timestamps are neither consumed by the simulator nor stored in the HDF5
payload. Accordingly, ``original-rate'' means that no common 25- or 240-fps
conversion is applied and that nominal/average source rates may differ across
videos; it does not imply preservation of per-frame PTS.

\begin{table}[t]
\centering
\caption{\textbf{Default parameters of the finalized video-to-voxel generator.}
Randomized quantities use the device-specific PyTorch generator described in
the text.}
\label{tab:v2v_parameters}
\small
\renewcommand{\arraystretch}{1.05}
\resizebox{\columnwidth}{!}{
\begin{tabular}{ll}
\toprule
\textbf{Parameter} & \textbf{Value / behavior} \\
\midrule
Decoded input & uint8 grayscale, source resolution \\
Reverse-gamma exponent & $2.2$ \\
Log-intensity offset & $0.001$ \\
Positive / negative threshold & $0.2 / 0.2$ (fixed) \\
Initial residual & $\mathcal{U}[-0.2,0.2)$ \\
Base noise & $\mathcal{N}(0,0.1^2)$ per transition/pixel \\
Hot-pixel model & Bernoulli $0.001$; amplitude $\mathcal{N}(0,0.1^2)$ \\
RNG seed & 42, reset per simulator call \\
Bins / frames per bin & $B\in\{5,9\}$ / $F=1$ \\
Temporal interpolation & None \\
Spatial transform & None \\
Output & int32 accumulator, clip, int16 HDF5 \\
Normalization & None \\
Refractory / leakage / shot noise & Absent \\
Per-frame PTS use & Absent; decoded frame order only \\
\bottomrule
\end{tabular}}
\end{table}

\begin{algorithm}[t]
\caption{Direct discrete video-to-voxel generation}
\label{alg:eventactivitynet-v2v}
\small
\begin{algorithmic}[1]
\Require Decoded grayscale frames $\{G_n\}_{n=0}^{N-1}$; bins
$B\in\{5,9\}$; frames per bin $F=1$
\State Set $c^+=c^-=0.2$, $\sigma_{\mathrm{base}}=0.1$,
$\rho_{\mathrm{hot}}=0.001$, and $\sigma_{\mathrm{hot}}=0.1$
\State Initialize $A\gets\mathbf{0}\in\mathbb{Z}^{B\times H\times W}$,
$t\gets0$, $b\gets0$, and residual $p\gets\varnothing$
\For{each resolution-dependent decode batch with its boundary frame}
  \State Reset the device RNG to seed 42
  \If{$p=\varnothing$}
    \State Sample $p(x)\sim\mathcal{U}[-c^-,c^+)$
  \EndIf
  \State Compute $L_n(x)=\log(0.001+\operatorname{clamp}
  (G_n(x)/255,0,1)^{2.2})$
  \State Sample a batch-local hot-pixel map $h(x)$
  \For{each adjacent-frame transition $i$ in the batch}
    \State Sample $\epsilon_i(x)\sim\mathcal{N}(0,\sigma_{\mathrm{base}}^2)$
    \State Compute $\delta_i$, cumulative state $s_i$, and levels
    $K_i^+,K_i^-$ using \Cref{eq:supp_v2v_state,eq:supp_v2v_levels}
    \State $q_i\gets(K_i^+-K_{i-1}^+)-(K_i^--K_{i-1}^-)$
    \State Store $q_i$ in the next bin of $A$
    \If{$b=B$}
      \State Clip $A$ to int16 and append one HDF5 tensor
      \State Record its transition start and count; reset $A$ and $b$
    \EndIf
  \EndFor
  \State Reduce the final state modulo $c^+$ or $c^-$ and carry it as $p$
\EndFor
\If{the final tensor is incomplete}
  \State Zero-pad trailing bins, clip to int16, and append it
\EndIf
\State \Return \texttt{events}, \texttt{voxel\_event\_start}, and
\texttt{voxel\_event\_count}
\end{algorithmic}
\end{algorithm}

\subsection{Held-Out Generator Sensitivity}
\label{secA:sim_compare}

We conduct a matched held-out comparison among
V2CE~\cite{ZhangCCYDMR24},
DVS-Voltmeter~\cite{LinMGW22}, and
V2V~\cite{abs-2505-16797}, the generator used for the
final Event ActivityNet release. All methods are evaluated
on the same source clips, spatial resolutions, and temporal
windows. For V2CE and DVS-Voltmeter, the generated
continuous events are accumulated into non-interpolated
5-bin voxel grids using the same frame-interval boundaries.
V2V directly produces the corresponding discrete 5-bin
voxels from consecutive source frames without materializing
a continuous event stream.

For each generator, the resulting voxel sequence is processed
by the same fixed HyperE2VID reconstructor, and reconstructed
frames are compared with temporally aligned source RGB
frames using MSE, SSIM, LPIPS-Alex, and LPIPS-VGG.
This comparison measures reconstruction-based content
retention under a common downstream pipeline; it does not
measure native-event realism or the fidelity of simulated
microsecond timestamps.

As shown in~\Cref{tab:sim_compare}, generator choice materially affects
reconstruction behavior. V2V obtains the lowest MSE and LPIPS-Alex and the
highest SSIM; its LPIPS-VGG is within $0.0006$ of DVS-Voltmeter. We adopt V2V
for the finalized payload because it directly produces the released discrete
voxel representation, avoids high-frame-rate interpolation and intermediate
continuous-event storage, and scales efficiently to the full 106.94-hour
collection. This engineering choice should not be interpreted as evidence
that V2V reproduces native sensor statistics more faithfully than
continuous simulators.

\begin{table}[t]
\centering
\caption{\textbf{Matched held-out generator sensitivity.}
V2CE and DVS-Voltmeter outputs are accumulated into
non-interpolated 5-bin voxels, whereas V2V directly
generates discrete 5-bin voxels. All rows use the same
source clips, temporal windows, spatial resolution, fixed
reconstructor, and evaluation implementation. Lower is
better for MSE and LPIPS; higher is better for SSIM.}
\label{tab:sim_compare}
\small
\renewcommand{\arraystretch}{1.08}
\resizebox{\columnwidth}{!}{
\begin{tabular}{lcccc}
\toprule
\textbf{Generator}
& \textbf{MSE ($\downarrow$)}
& \textbf{SSIM ($\uparrow$)}
& \textbf{LPIPS-Alex ($\downarrow$)}
& \textbf{LPIPS-VGG ($\downarrow$)} \\
\midrule
V2CE~\cite{ZhangCCYDMR24}
& $0.0708 \pm 0.0433$
& $0.4315 \pm 0.1363$
& $0.5114 \pm 0.1007$
& $0.5817 \pm 0.0583$ \\

DVS-Voltmeter~\cite{LinMGW22}
& $0.0643 \pm 0.0361$
& $0.5920 \pm 0.1099$
& $0.3854 \pm 0.1163$
& $\mathbf{0.4767 \pm 0.0742}$ \\

V2V~\cite{abs-2505-16797}
& $\mathbf{0.0623 \pm 0.0321}$
& $\mathbf{0.6104 \pm 0.0152}$
& $\mathbf{0.3714 \pm 0.1057}$
& $0.4773 \pm 0.0642$ \\
\bottomrule
\end{tabular}
}
\end{table}

\subsection{Caption--Action Alignment Quality}
\label{secA:caption_quality}

ActivityNet Captions and ActivityNet v1.3 were created with different temporal segmentation and annotation goals. We therefore treat caption matches as weak auxiliary supervision rather than hard action labels. Each caption segment is assigned to the ActivityNet v1.3 action instance with maximum temporal IoU, and matches below $\eta=0.1$ are excluded from the alignment objective.

Among 15{,}653 caption segments, 9{,}527 (60.86\%) satisfy the matching threshold, while 6{,}126 (39.14\%) are marked unmatched. The retained matches include 3{,}577 captions in $[0.1,0.3)$, 2{,}477 in $[0.3,0.5)$, 1{,}645 in $[0.5,0.7)$, and 1{,}828 in $[0.7,1.0]$. Thus, 37.55\% of retained captions lie in the low-overlap interval $[0.1,0.3)$. This distribution reflects the different annotation purposes of ActivityNet Captions and ActivityNet v1.3 and motivates treating captions as weak rather than hard labels. Caption matches neither redefine recognition labels nor modify the original TAL boundaries. Because temporal overlap alone does not guarantee semantic equivalence, we do not infer caption usefulness from these statistics; instead,~\Cref{tab:caption_supervision_ablation} directly compares prompt-only, caption-only, combined, and threshold-controlled variants.

\begin{figure*}[t]
    \centering
    \includegraphics[width=0.78\textwidth]{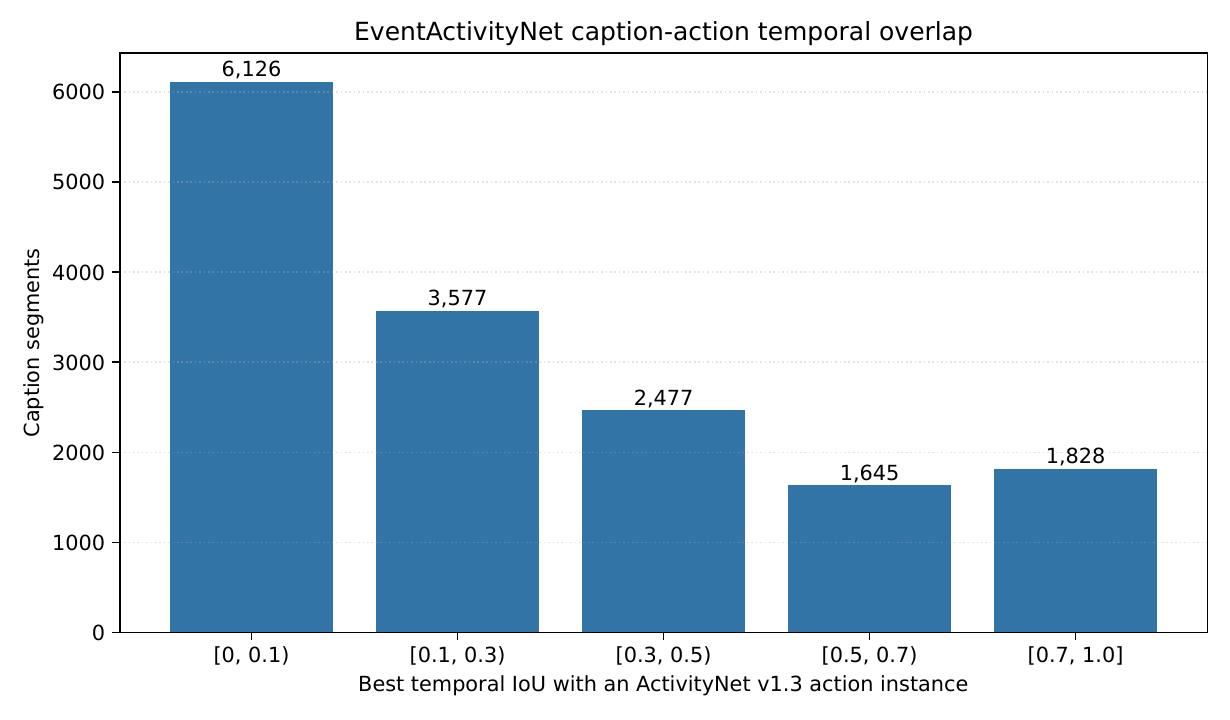}
    \caption{\textbf{Caption--action temporal-alignment distribution.}
    We show the best temporal IoU between each ActivityNet Captions segment and an ActivityNet v1.3 action instance. At $\eta=0.1$, 9{,}527 of 15{,}653 captions are retained for weak event--language supervision and 6{,}126 are excluded. Among retained captions, 3{,}577 fall in the low-overlap interval $[0.1,0.3)$, highlighting that temporal overlap does not imply exact semantic correspondence.}
    \label{fig:caption_alignment}
\end{figure*}

\subsection{Controlled Caption-Supervision Ablation}
\label{secA:caption_ablation}

We isolate the contribution of captions under a matched Event ActivityNet
(Small) recognition protocol. All rows use the 5-bin input, AEF-SM,
$1{\times}1$ channel projection, ExACT backbone, frozen text encoder,
supervised recognition loss, fixed 100-epoch schedule, final-epoch reporting rule,
and random-seed policy used by the full contrastive model. The only change is
the auxiliary text-positive set. Class-prompt-only supervision removes all
caption terms; caption-only supervision removes the class-prototype terms
from the alignment branch; and the combined setting uses both. The threshold
sweep keeps the best-tIoU assignment fixed and changes only which captions
are eligible positives.

The retained-positive counts follow the fixed alignment manifest:
$\eta=0.1$ retains 9{,}527 captions, $\eta=0.3$ retains 5{,}950, and
$\eta=0.5$ retains 3{,}473. These counts are obtained from the same
15{,}653-caption pool and do not change the video split or action labels.

\begin{table}[t]
\centering
\caption{\textbf{Controlled contribution of class prompts and matched
captions on Event ActivityNet (Small).}
All rows use the same recognition model, supervised classification loss,
training budget, final-epoch reporting rule, and evaluator. ``Caption positives''
reports the number of caption segments eligible under the indicated
best-tIoU threshold.}
\label{tab:caption_supervision_ablation}
\small
\renewcommand{\arraystretch}{1.06}
\resizebox{\columnwidth}{!}{
\begin{tabular}{lccccl}
\toprule
\textbf{Auxiliary alignment} &
\textbf{$\eta$} &
\textbf{Caption positives} &
\textbf{Top-1} &
\textbf{Top-5} &
\textbf{Purpose} \\
\midrule
None
& -- & 0 & 53.31 & 57.93
& No text alignment \\
Class prompts only
& -- & 0 & 56.10 & 60.15
& Class semantics \\
Captions only
& 0.1 & 9{,}527 & 54.85 & 59.25
& Instance descriptions \\
Class prompts + captions
& 0.1 & 9{,}527 & 57.24 & 61.35
& Full setting \\
Class prompts + captions
& 0.3 & 5{,}950 & \textbf{57.65} & \textbf{61.75}
& Remove low-overlap positives \\
Class prompts + captions
& 0.5 & 3{,}473 & 57.10 & 61.20
& Higher-confidence positives \\
\bottomrule
\end{tabular}}
\end{table}

The comparisons answer distinct questions. Class prompts only versus no
alignment measures the value of category-level text. Captions only versus no
alignment measures whether instance descriptions help without prompt
prototypes. The full setting versus class prompts only measures the
incremental contribution of captions, while the $\eta$ sweep tests whether
that contribution is robust to removing the 3{,}577 low-overlap captions in
$[0.1,0.3)$. Captions alone improve the no-alignment baseline from 53.31/57.93 to 54.85/59.25, and adding captions to class prompts raises 56.10/60.15 to 57.24/61.35 at $\eta=0.1$. Removing low-overlap positives at $\eta=0.3$ gives the best result, 57.65/61.75, whereas the stricter $\eta=0.5$ setting remains close at 57.10/61.20. Thus, captions provide a modest complementary gain, and that gain is not driven solely by the lowest-overlap matches.

\subsection{Event--Text Objective Details}
\label{secA:event_text_objective}

\subsubsection{Event and Text Representations.}
An event encoder $f_{\theta}$ maps $X_i$ to
$\mathbf{z}_i=\operatorname{norm}(f_{\theta}(X_i))$. For class $y_i$, the
frozen text encoder $g$ produces handcrafted-prompt, learnable-prompt, and
matched-caption embeddings. The handcrafted template is ``A series of photos
recording action for [CLASS].'', while the learnable template prepends $M$
learnable tokens to the class name. Denoting these templates by $\pi_h$ and
$\pi_\ell$, respectively,
\begin{equation}
\begin{aligned}
\mathbf{t}_{y_i}^{(h)}
&=\operatorname{norm}\!\left(g(\pi_h(y_i))\right),\\
\mathbf{t}_{y_i}^{(\ell)}
&=\operatorname{norm}\!\left(g(\pi_\ell(y_i))\right),\\
\mathbf{t}_{i}^{(\mathrm{cap})}
&=\operatorname{norm}\!\left(g(\mathrm{caption}_i)\right).
\end{aligned}
\label{eq:supp_event_text_embeddings}
\end{equation}
The class prototype averages the two prompt branches and is normalized again:
\begin{equation}
\mathbf{p}_{y_i}
=
\operatorname{norm}\!\left(
\frac{\mathbf{t}_{y_i}^{(h)}+\mathbf{t}_{y_i}^{(\ell)}}{2}
\right).
\label{eq:supp_class_prototype}
\end{equation}
An optional text projection maps all text features to the event-embedding
dimension. Similarity is temperature-scaled cosine similarity,
$s(\mathbf{a},\mathbf{b})=\mathbf{a}^{\top}\mathbf{b}/\tau$.

\subsubsection{Contrastive Event--Text Alignment.}
Let $\mathcal{B}$ be a mini-batch, $n_{\mathrm b}=|\mathcal{B}|$,
$m_i\in\{0,1\}$ indicate whether sample $i$ has an eligible caption, and
$\mathcal{C}_{\mathcal B}$ be the unique classes in the mini-batch. The
positive mass for anchor $\mathbf z_i$ is
\begin{equation}
P_i=
\exp s(\mathbf{z}_i,\mathbf{p}_{y_i})
+
m_i\exp s(\mathbf{z}_i,\mathbf{t}_{i}^{(\mathrm{cap})}),
\label{eq:supp_event_text_positive}
\end{equation}
and the normalization term is
\begin{equation}
\begin{aligned}
Z_i
={}&
\sum_{c\in\mathcal{C}_{\mathcal B}}
\exp s(\mathbf{z}_i,\mathbf{p}_{c})
+
m_i\exp s(\mathbf{z}_i,\mathbf{t}_{i}^{(\mathrm{cap})})\\
&+
\lambda_{\mathrm{cap}}
\sum_{\substack{j\ne i\\m_j=1}}
\exp s(\mathbf{z}_i,\mathbf{t}_{j}^{(\mathrm{cap})}).
\end{aligned}
\label{eq:supp_event_text_normalizer}
\end{equation}
The complete auxiliary objective is
\begin{equation}
\mathcal{L}_{\mathrm{con}}
=
-\frac{1}{n_{\mathrm b}}
\sum_{i=1}^{n_{\mathrm b}}
\log\frac{P_i}{Z_i}.
\label{eq:supp_lalign_ecp}
\end{equation}
Only caption negatives are down-weighted by
$\lambda_{\mathrm{cap}}$, because captions may describe context or
co-occurring actions. Class prototypes remain fully weighted, and unmatched
captions are excluded through $m_i$.

\subsubsection{Controlled Caption-Supervision Variants.}
The class-prompt-only variant removes all caption terms from $P_i$ and $Z_i$.
The caption-only variant removes class-prototype terms from the auxiliary
alignment loss; samples without an eligible caption still contribute to the
unchanged supervised recognition loss. The combined variant uses
\Cref{eq:supp_event_text_positive,eq:supp_event_text_normalizer} as written.
For threshold $\eta$, the best-tIoU assignment is fixed and only the
eligibility mask changes:
\begin{equation}
m_i=\mathbb{I}\!\left[
\max_k\operatorname{tIoU}(C_i,A_k)\ge\eta
\right].
\label{eq:supp_caption_eligibility}
\end{equation}
Changing $\eta$ therefore does not alter action labels, temporal boundaries,
event inputs, the frozen text encoder, or the training budget.

\subsubsection{Positive-Only Smooth-$\ell_1$ Alignment.}
Following the positive-only feature-regression design studied by
RECON~\cite{qi2023recon}, let $\ell_{\mathrm{SL1}}$ denote Smooth-$\ell_1$
loss and $\operatorname{sg}$ stop-gradient. We evaluate
\begin{equation}
\begin{aligned}
\mathcal{L}_{\mathrm{pos}}
=
\frac{1}{n_{\mathrm b}}\sum_{i=1}^{n_{\mathrm b}}
\Big[
&\lambda_{\mathrm{cls}}\,
\ell_{\mathrm{SL1}}\!\left(
\mathbf{z}_i,\operatorname{sg}(\mathbf{p}_{y_i})
\right)\\
&+
m_i\lambda_{\mathrm{cap}}\,
\ell_{\mathrm{SL1}}\!\left(
\mathbf{z}_i,\operatorname{sg}(\mathbf{t}_{i}^{(\mathrm{cap})})
\right)
\Big].
\end{aligned}
\label{eq:supp_lalign_positive}
\end{equation}
This is an ablation rather than the primary objective. All variants retain the
same supervised recognition loss; only the auxiliary text-alignment terms
change. The controlled results are reported in
\Cref{tab:caption_supervision_ablation}.

\subsection{LPIPS Configuration and Sensitivity}
\label{secA:lpips_sensitivity}

\subsubsection{Reconstructor and Frame Mapping.}
We use the publicly released pretrained HyperE2VID model~\cite{ercan2024hypere2vid} with its default inference configuration and without task-specific fine-tuning. For each annotated action segment $(s,e)$, the midpoint $t_c=(s+e)/2$ is mapped to the corresponding source frame and 5-bin voxel index through the released variable-FPS timing metadata. A fixed-length voxel window centered at that index is passed to HyperE2VID, and only the center reconstruction is retained. This reconstructed frame is compared with the source RGB frame aligned to the same midpoint. Consequently, every annotated action instance contributes one perceptual comparison, independent of its duration.

\subsubsection{LPIPS Configuration.}
Perceptual distance is computed with the official LPIPS implementation~\cite{zhang2018unreasonable}. We use its default calibrated AlexNet backbone and evaluate RGB tensors normalized to $[-1,1]$. No gradients are propagated through HyperE2VID or LPIPS. For a video $V$, we aggregate segment-level scores using
\begin{equation}
q(V)=\max_{(s,e)\in\mathcal{S}(V)}
\operatorname{LPIPS}\!\left(I_{t_c},\hat I_{t_c}\right).
\label{eq:supp_lpips_video_score}
\end{equation}
so a video is marked as a high-LPIPS case whenever at least one annotated
action center exceeds the reference threshold. The threshold is used for
release auditing rather than automatic exclusion, and all 3{,}263
structurally validated videos remain in the benchmark. Because only action
centers are evaluated, this diagnostic does not fully assess temporal
boundaries or background-to-action transitions.

\subsubsection{Threshold Sensitivity.}
\Cref{tab:lpips_threshold} partitions all 3{,}263 finalized videos by
whether $q(V)$ is above or at/below each threshold. Increasing $\delta$ can
only reduce the above-threshold set. We use $\delta=0.30$ as the reference
diagnostic threshold, not as an exclusion rule; neighboring thresholds show
the sensitivity of this soft quality-control flag.

\begin{table}[t]
\centering
\caption{\textbf{Sensitivity of the action-center LPIPS diagnostic.}
We report the number of finalized videos whose maximum action-center score is
above or at/below each threshold. No threshold changes the 3{,}263-video
release.}
\label{tab:lpips_threshold}
\small
\renewcommand{\arraystretch}{1.08}
\begin{tabular}{cccc}
\toprule
\textbf{$\delta$} &
\textbf{Above} &
\textbf{At/Below} &
\textbf{Above-threshold Rate} \\
\midrule
0.15 & 1{,}035 & 2{,}228 & 31.72\% \\
0.20 & 610 & 2{,}653 & 18.69\% \\
0.25 & 209 & 3{,}054 & 6.41\% \\
0.30 & 0 & 3{,}263 & 0.00\% \\
\bottomrule
\end{tabular}
\end{table}

\subsection{Post-Generation Event Diagnostics}
\label{secA:event_diagnostics}

Reconstruction-based LPIPS measures whether a fixed reconstructor can recover source content around annotated action centers, but it does not characterize the numerical behavior of the released voxel tensors. We therefore conduct a bounded, sample-based post-generation diagnostic on the finalized 5-bin release. These measurements are \emph{voxel activity statistics}; they are not exact counts of native asynchronous events and do not establish equivalence to a physical sensor.

\subsubsection{Sampling and Validation Protocol.}
We deterministically select 200 videos with seed 2025, covering all 200 action classes. The sample contains 142 training and 58 validation videos, with 66 short, 66 medium, and 68 long videos. For each video, we uniformly sample at most 64 temporal indices over the complete voxel sequence and read each tensor independently. The sampled-time denominator follows the released per-video variable-FPS timing metadata. All 200 HDF5 files are readable and satisfy the expected schema. We observe no malformed files, NaN/Inf values, int16 saturation, or values with absolute magnitude at least 30{,}000.

For sampled voxel tensor $X_t\in\mathbb{R}^{B\times H\times W}$, normalized activity mass is the accumulated absolute voxel mass divided by sampled duration and spatial resolution. Positive-mass ratio measures the fraction of absolute mass with positive polarity. Active-cell ratio is the proportion of nonzero tensor cells. Let
\begin{equation}
\begin{aligned}
M_t(x,y)&=\sum_b |X_t[b,x,y]|,\\
q_t(x,y)&=\frac{M_t(x,y)}{\sum_{x,y}M_t(x,y)}.
\end{aligned}
\label{eq:supp_voxel_mass_distribution}
\end{equation}
We define normalized spatial entropy as
\begin{equation}
H_{\mathrm{spatial}}(t)
=
-\frac{\sum_{x,y}q_t(x,y)\log q_t(x,y)}
{\log(HW)}.
\end{equation}
Temporal activity variation is summarized by the coefficient of variation of per-index absolute voxel mass.

\begin{table}[t]
\centering
\caption{\textbf{Sample-based post-generation statistics for the finalized 5-bin release.}
The deterministic 200-video sample covers all 200 classes and the full split/duration structure. P05 and P95 are computed over per-video statistics. Values describe voxel activity and should not be interpreted as native-event counts.}
\label{tab:event_diagnostics_5bin}
\small
\renewcommand{\arraystretch}{1.08}
\resizebox{\columnwidth}{!}{
\begin{tabular}{lcccc}
\toprule
\textbf{Metric} & \textbf{Mean} & \textbf{Median} & \textbf{P05} & \textbf{P95} \\
\midrule
Normalized voxel activity mass & 30.1649 & 27.6494 & 14.0468 & 53.7894 \\
Positive-mass ratio & 0.500051 & 0.500315 & 0.475312 & 0.519415 \\
Active-cell ratio & 0.510680 & 0.499737 & 0.404583 & 0.658778 \\
Normalized spatial entropy & 0.969674 & 0.970523 & 0.955633 & 0.979605 \\
Temporal activity CV & 0.510191 & 0.495831 & 0.221550 & 0.867798 \\
Absolute-value P95 & 4.190 & 4.000 & 1.000 & 8.000 \\
Absolute-value P99 & 10.805 & 10.000 & 3.950 & 19.050 \\
Maximum absolute value & 34.340 & 35.000 & 29.000 & 37.000 \\
\bottomrule
\end{tabular}
}
\end{table}

The polarity mass is nearly balanced around 0.5, while the active-cell ratio and temporal coefficient of variation show substantial cross-video variation. Spatial entropy is consistently high under this normalized statistic, indicating broadly distributed activity in the sampled tensors rather than concentration in a small number of pixels. The maximum observed magnitudes remain far from the int16 limits, and the absence of saturation or extreme values provides an additional numerical-integrity check. These results support the use of the release for downstream voxel-based experiments, but they do not measure sensor realism or close the synthetic-to-native gap.

\subsection{Subset Curation and Distribution Shift}
\label{secA:subset_stats}

This subsection reports the composition of the curated subset, the
motion/illumination enrichment signals, and the distribution shift relative
to the ActivityNet Captions source pool.

\subsubsection{Selection and Composition.}
\begin{figure*}[t]
    \centering
    \includegraphics[width=0.97\textwidth]{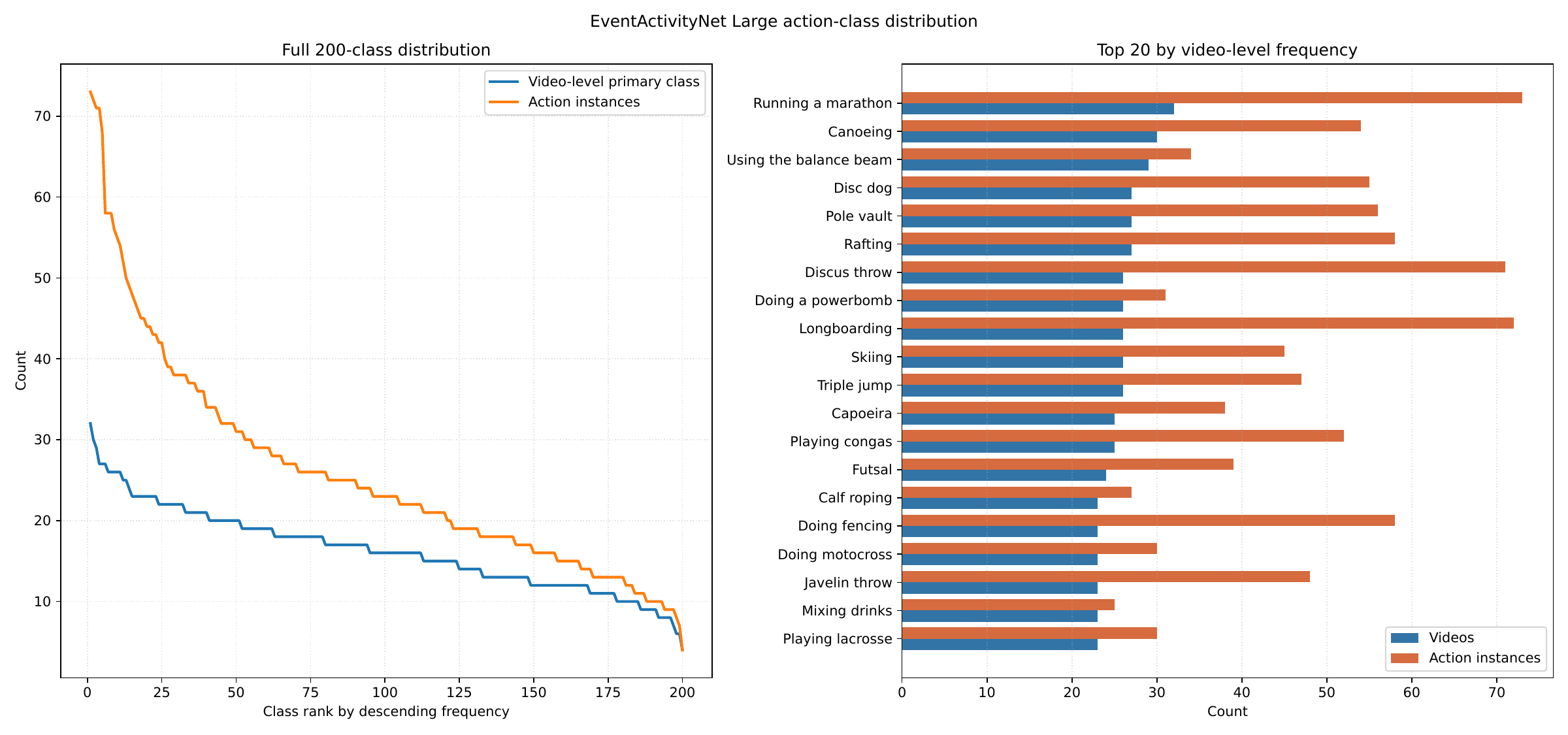}
    \caption{\textbf{Class distribution of the finalized subset.}
    Left: video-level class counts sorted by frequency.
    Right: the 20 most frequent classes.
    The 3{,}263-video subset retains all 200 ActivityNet action classes while preserving a long-tailed distribution.}
    \label{fig:s1}
\end{figure*}

The source pool comprises 14{,}926 ActivityNet Captions v1.3
train/validation videos~\cite{KrishnaHRFN17}. The finalized Event ActivityNet
subset contains 3{,}263 videos and 106.94 hours from all 200 primary action
classes, with 2{,}316 training and 947 validation videos. Its duration strata
contain 1{,}077 short, 1{,}076 middle-duration, and 1{,}110 long videos.
\Cref{fig:s1} shows the retained video-level class distribution, and
\Cref{fig:s2}(a) summarizes the duration composition.

\subsubsection{Motion/Illumination Enrichment Signals.}
\begin{figure*}[t]
    \centering
    \includegraphics[width=0.97\textwidth]{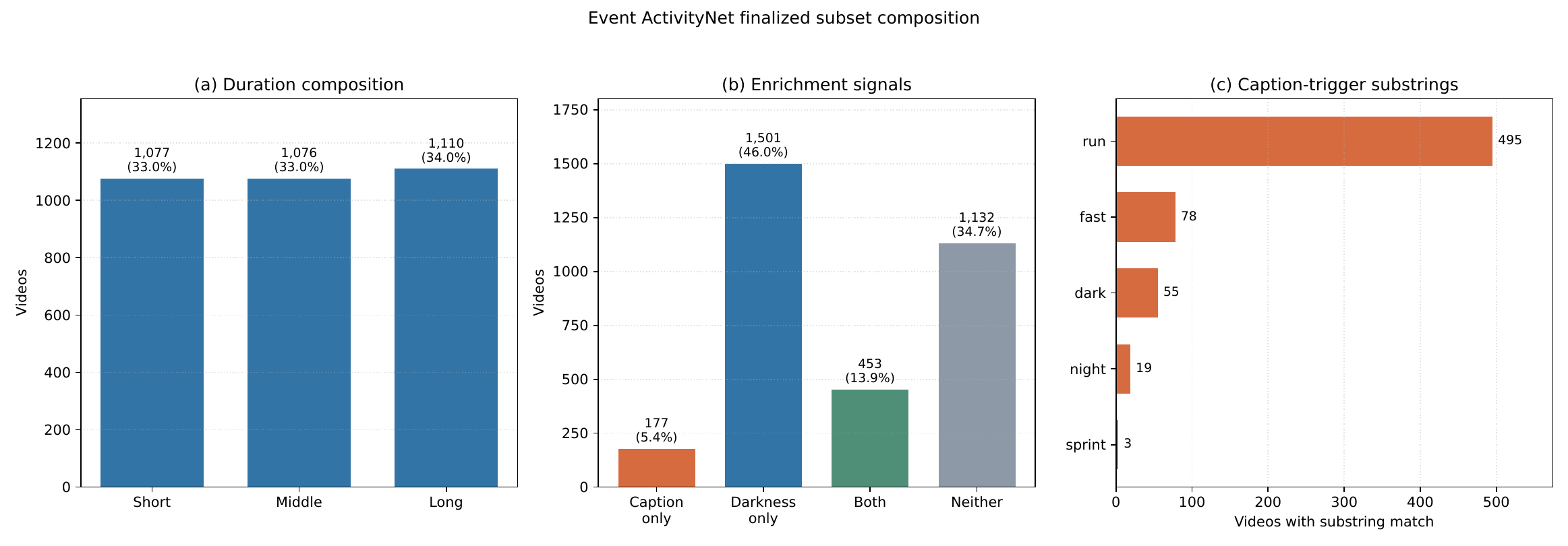}
    \caption{\textbf{Duration structure and enrichment signals in the curated subset.}
    (a) The finalized 3{,}263-video subset contains 1{,}077 short,
    1{,}076 middle-duration, and 1{,}110 long videos.
    (b) Among the finalized videos, 177 satisfy only the caption-trigger
    criterion, 1{,}501 only the first-frame darkness criterion, 453 satisfy
    both, and 1{,}132 satisfy neither.
    (c) Video-level frequencies of the nonzero caption-trigger substrings
    observed in the finalized subset. A video may match multiple substrings.}
    \label{fig:s2}
\end{figure*}

In the finalized subset, 630 videos have a caption trigger, 1{,}954 have a
first-frame darkness trigger, and 2{,}131 (65.31\%) satisfy at least one of
the two signals. Their joint composition is 177 caption-trigger-only videos,
1{,}501 darkness-trigger-only videos, 453 videos satisfying both signals, and
1{,}132 satisfying neither. The nonzero caption-trigger substrings observed
in the finalized subset occur in 495 videos for \emph{run}, 78 for
\emph{fast}, 55 for \emph{dark}, 19 for \emph{night}, and 3 for
\emph{sprint}. These video-level counts overlap because one video may match
multiple substrings. \Cref{fig:s2}(b--c) visualizes the mutually exclusive
trigger composition and the overlapping substring frequencies.

\subsubsection{Class-Distribution Shift.}
\begin{figure*}[t!]
    \centering
    \includegraphics[width=0.97\textwidth]{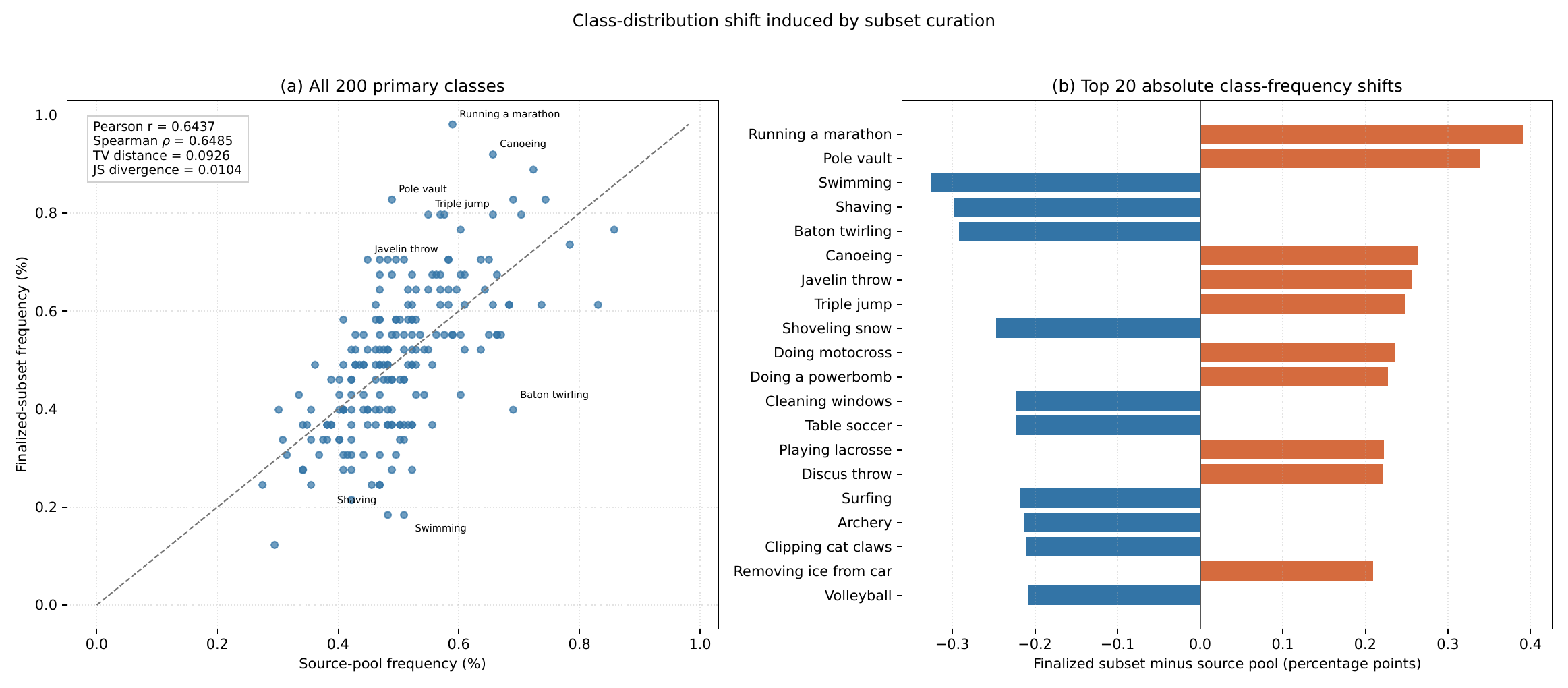}
    \caption{\textbf{Class-distribution shift induced by subset curation.}
    We compare normalized primary-class frequencies between the
    14{,}926-video source pool and the finalized 3{,}263-video subset.
    All 200 classes are retained, with Pearson correlation $r=0.6437$,
    Spearman correlation $\rho=0.6485$, total-variation distance $0.0926$,
    and Jensen--Shannon divergence $0.0104$.
    The remaining shifts reflect class-coverage constraints, duration
    stratification, and motion/illumination-oriented enrichment.}
    \label{fig:s3}
\end{figure*}

We quantify class-distribution shift using one primary action class per video
and separately normalized 200-class frequency vectors for the source pool and
finalized subset. Their Pearson class-frequency correlation is $r=0.6437$,
and their Spearman rank correlation is $\rho=0.6485$. The total-variation
distance is $0.0926$, and the base-2 Jensen--Shannon divergence is $0.0104$.
All 200 source-pool classes remain represented, while the nonzero differences
reflect class-coverage constraints, duration stratification, and
motion/illumination-oriented enrichment. \Cref{fig:s3} shows the full
source-to-subset comparison and the 20 largest absolute class-frequency
shifts.

\subsection{Release Packaging and Integrity}
\label{secA:release_integrity}

The matched 5-bin and 9-bin releases use identical video IDs, inherited
annotations, timing records, split definitions, and ID manifests. Both are
distributed as sharded train/validation archives and each occupies
approximately 5~TB. The verified 5-bin payload contains 3{,}263 HDF5 samples
in 157 training and 62 validation tar shards (219 shards total), with
219/219 shard-level SHA256 checksums verified.

Each HDF5 sample stores the int16 voxel tensor in \texttt{events}, together
with \texttt{voxel\_event\_start} and \texttt{voxel\_event\_count} indexing
the transition range represented by each voxel. Shared metadata include
3{,}263 rational-rate and duration timing records, ActivityNet Captions,
ActivityNet v1.3 action instances, caption--action alignment files,
annotation-issue records, dataset documentation, and fixed nested-scale and
train/validation manifests. The 9-bin payload follows the same organization.

Release validation checks HDF5 readability, tensor/index consistency, timing
metadata, split membership, manifest membership, annotation records, shard
membership, and checksums. These checks establish payload and metadata
integrity; they are separate from the reconstruction-based LPIPS diagnostic,
which assesses retained reconstructable content at annotated action centers.

\subsection{Event-Based Action Recognition}
\label{secA:event_ar}

\subsubsection{Window Statistics.}
For an interval $W=[t_s,t_e)$, let $\mathcal{Q}(W)$ denote the voxel indices
whose timestamps fall inside $W$. We summarize the interval by
\begin{equation}
\begin{aligned}
\phi(W)&=[\rho(W),\mathbf h(W)],\\
\rho(W)&=
\frac{\sum_{q\in\mathcal Q(W)}\lVert X_q\rVert_1}{t_e-t_s},\\
\mathbf h(W)&=
\operatorname{Norm}_1\!\left(
\sum_{q\in\mathcal Q(W)}\operatorname{Hist}_K(X_q)
\right),
\end{aligned}
\label{eq:supp_AEF_stats}
\end{equation}
where $\rho(W)$ is voxel activity per unit time and
$\operatorname{Hist}_K$ accumulates polarity-aware voxel mass into $K$ coarse
spatial-cell--polarity bins. Thus, $\mathbf h(W)$ captures activity
location and polarity, while $\rho(W)$ captures magnitude.

\subsubsection{Window Dissimilarity.}
Two windows are compared by
\begin{equation}
\begin{aligned}
D_{\boldsymbol\omega}(W_a,W_b)
={}&
\omega_\rho
\left|
\log\frac{\rho(W_a)+\epsilon}
         {\rho(W_b)+\epsilon}
\right|\\
&+
\omega_h D_{\mathrm{JS}}\!\left(
\mathbf h(W_a),\mathbf h(W_b)
\right).
\end{aligned}
\label{eq:supp_AEF_distance}
\end{equation}
where $\epsilon$ stabilizes the log ratio and
$\boldsymbol\omega=(\omega_\rho,\omega_h)$ balances activity-rate and
distributional changes.

\subsubsection{Adaptive Splitting.}
For a candidate pivot $t_m$ in $W=[t_s,t_e)$, define
$W_L=[t_s,t_m)$ and $W_R=[t_m,t_e)$. The split score and selected pivot are
\begin{equation}
\begin{aligned}
S_{\mathrm{split}}(W,t_m)
&=
D_{\boldsymbol\omega}(W_L,W_R),\\
t_m^\star
&=
\arg\max_{t_m}
S_{\mathrm{split}}(W,t_m).
\end{aligned}
\label{eq:supp_AEF_split}
\end{equation}
The window is split when
$S_{\mathrm{split}}(W,t_m^\star)>\tau_{\mathrm{split}}$, subject to
$(t_m^\star-t_s)\ge d_{\min}$ and
$(t_e-t_m^\star)\ge d_{\min}$.

\subsubsection{Adaptive Merging and Execution Order.}
For adjacent windows $W_i$ and $W_{i+1}$,
\begin{equation}
S_{\mathrm{merge}}(W_i,W_{i+1})
=
D_{\boldsymbol\omega}(W_i,W_{i+1}).
\label{eq:supp_AEF_merge}
\end{equation}
and they are merged when
$S_{\mathrm{merge}}(W_i,W_{i+1})<\tau_{\mathrm{merge}}$, optionally subject
to a maximum-duration constraint. AEF-SM starts from a coarse fixed-length
partition, applies splitting to convergence, and then performs a greedy merge
pass until no adjacent pair satisfies the criterion. The minimum duration
prevents fragmented windows, the high split threshold isolates pronounced
internal changes, and the lower merge threshold removes only strongly
consistent neighbors.

\subsubsection{Default AEF-SM Parameters.} 
Unless specified, we use $d_{\min}=0.5$s, $K=8\times 8\times 2$ histogram bins,
$\omega_\rho=\omega_h=1$, and pivot candidates every 0.25s.
We set $\tau_{\rm split}$ as the 95th percentile of $S_{\rm split}$ on the training set,
and $\tau_{\rm merge}$ as the 20th percentile of $S_{\rm merge}$.

To prevent split--merge oscillation, we apply splitting to convergence and then perform a merge pass (\Cref{alg:AEF}), without interleaving.
We enforce hysteresis by using \(\tau_{\mathrm{merge}} < \tau_{\mathrm{split}}\) (default: 20th vs.\ 95th percentile).

\subsection{Public Event-Recognition Evaluations}
\label{secA:public_recognition}

We additionally evaluate the recognition implementation on PAF, HARDVS,
DVS128 Gesture, and SeAct following each public benchmark's ExACT protocol.
These experiments verify the baseline interfaces but are not pooled with
Event ActivityNet and do not define a cross-dataset ranking.

\begin{table}[t]
\centering
\caption{\textbf{Independent public event-recognition evaluations.}
Entries are Top-1/Top-5 accuracy under each benchmark's standard ExACT
protocol.}
\label{tab:supp_public_recognition}
\small
\renewcommand{\arraystretch}{1.05}
\resizebox{\columnwidth}{!}{
\begin{tabular}{lcc}
\toprule
\textbf{Dataset / Setting} & \textbf{Top-1} (\%) & \textbf{Top-5} (\%) \\
\midrule
PAF (ExACT-category)~\cite{MiaoCNZRBK19} & 94.83 & 98.28 \\
HARDVS (ExACT-category)~\cite{wang2024hardvs} & 90.10 & 96.69 \\
DVS128 Gesture (ExACT-category)~\cite{amir2017low} & 98.86 & 98.86 \\
SeAct (ExACT-category)~\cite{ZhouZLW24} & 66.07 & 70.54 \\
SeAct (ExACT-caption)~\cite{ZhouZLW24} & 67.24 & 75.00 \\
\bottomrule
\end{tabular}}
\end{table}

\begin{table}[t]
\centering
\caption{\textbf{Controlled AEF-SM evaluation on PAF.}
Both rows use the 100k-event public-data input, the same ExACT backbone,
$1{\times}1$ projection, schedule, and evaluation protocol; only AEF-SM
changes. This setting is distinct from the independent ExACT-category row in
\Cref{tab:supp_public_recognition}.}
\label{tab:supp_public_aef}
\small
\renewcommand{\arraystretch}{1.05}
\begin{tabular}{lcc}
\toprule
\textbf{Setting} & \textbf{Top-1} (\%) & \textbf{Top-5} (\%) \\
\midrule
PAF (100k), w/o AEF-SM & 90.85 & 92.31 \\
PAF (100k), w/ AEF-SM & \textbf{92.45} & \textbf{94.57} \\
\bottomrule
\end{tabular}
\end{table}

\begin{table}[t]
\centering
\caption{\textbf{Event--text alignment on public recognition datasets.}
We compare positive-only Smooth-$\ell_1$ regression with contrastive
alignment.}
\label{tab:supp_public_alignment}
\small
\renewcommand{\arraystretch}{1.05}
\resizebox{\columnwidth}{!}{
\begin{tabular}{lcc}
\toprule
\textbf{Dataset / Setting} & \textbf{Top-1} (\%) & \textbf{Top-5} (\%) \\
\midrule
HARDVS / ExACT-category (baseline) & 90.10 & 96.69 \\
HARDVS / ExACT-category + Positive-only & 89.02 & 94.25 \\
HARDVS / ExACT-category + Contrastive & \textbf{92.45} & \textbf{98.56} \\
\midrule
SeAct / ExACT-category (baseline) & 66.07 & 70.54 \\
SeAct / ExACT-category + Positive-only & 65.33 & 70.20 \\
SeAct / ExACT-category + Contrastive & \textbf{69.42} & \textbf{72.61} \\
\midrule
SeAct / ExACT-caption (baseline) & 67.24 & 75.00 \\
SeAct / ExACT-caption + Positive-only & 66.26 & 72.35 \\
SeAct / ExACT-caption + Contrastive & \textbf{69.14} & \textbf{76.79} \\
\bottomrule
\end{tabular}}
\end{table}

\subsection{Event-Based Online TAL}
\label{secA:event_ontal}

\subsubsection{Causal Feature Interface and Temporal Alignment.}
At step $t$, the selected localization head receives an RGB feature prefix
$\mathbf R_{\le t}$, an event feature prefix $\mathbf E_{\le t}$, or a fused
prefix derived from the same observed video prefix. Let
$\mathbf R\in\mathbb R^{T_r\times d_r}$ and
$\mathbf E\in\mathbb R^{T_e\times d_e}$. C1 downsamples events to the RGB
rate,
\begin{equation}
\tilde{\mathbf E}_t=
\operatorname{Agg}\!\left(\mathbf E_{\mathcal I(t)}\right),
\qquad
\tilde{\mathbf E}\in\mathbb R^{T_r\times d_e},
\label{eq:supp_align_downsample}
\end{equation}
where $\mathcal I(t)$ indexes event steps in the $t$-th RGB interval and
$\operatorname{Agg}$ is average or sum pooling. C2 replicates RGB features to
the event rate,
\begin{equation}
\tilde{\mathbf R}_{t,n}=\mathbf R_t,\quad n=1,\ldots,N,
\qquad
\tilde{\mathbf R}\in\mathbb R^{(T_rN)\times d_r}.
\label{eq:supp_align_upsample}
\end{equation}

\subsubsection{Input Variants.}
The matched RGB-only control aligns RGB features to the CA streaming
timestamps and applies the same projection used by the RGB branch:
\begin{equation}
\mathbf F_{t'}=
\operatorname{Proj}_{R}\!\left(\tilde{\mathbf R}_{t'}\right).
\label{eq:supp_rgb_only}
\end{equation}
The event-only replacement uses $\mathbf F=\mathbf E$. Concatenation uses
\begin{equation}
\mathbf F_t=[\mathbf R_t;\tilde{\mathbf E}_t]
\quad\text{or}\quad
\mathbf F_{t'}=[\tilde{\mathbf R}_{t'};\mathbf E_{t'}],
\label{eq:supp_fuse_concat}
\end{equation}
followed by a lightweight MLP or 1D convolution. Cross-attention conditions
the event prefix on RGB context:
\begin{equation}
\mathbf F_{\le t}
=
\operatorname{CA}\!\left(
\mathbf Q=\mathbf E_{\le t},
\mathbf K=\mathbf R_{\le t},
\mathbf V=\mathbf R_{\le t};
\mathbf M_{\mathrm{causal}}
\right).
\label{eq:supp_fuse_xattn}
\end{equation}
The causal mask prevents access to future RGB features. Past RGB features are
cached and the cache is updated monotonically. The same feature-construction
variants are used with SimOn and MATR; only the localization head changes.

\subsubsection{MATR Boundary Refinement.}
Let MATR~\cite{song2024online} produce a coarse proposal $(s_k^{c},e_k^{c},a_k^{c})$ from its memory-augmented encoder, separate start/end decoders, and prediction heads. For a proposal-aligned event or fused feature $\mathbf{u}_k$, lightweight residual heads predict
\begin{equation}
(\Delta s_k,\Delta e_k)=h_{\mathrm{off}}(\mathbf{u}_k).
\label{eq:matr_residual_offsets}
\end{equation}
The refined proposal is
\begin{equation}
\tilde{s}_k=s_k^{c}+\Delta s_k,\qquad
\tilde{e}_k=e_k^{c}+\Delta e_k,\qquad
\tilde{a}_k=a_k^{c}.
\label{eq:matr_refined_proposal}
\end{equation}
Thus, the reported offset-head setting refines temporal boundaries while retaining MATR's class prediction. The residual heads are trained with Smooth-$\ell_1$ boundary regression together with the original MATR classification and localization losses. All predictions remain prefix-only, and previously emitted instances are not modified.

\FloatBarrier
\section{Reference Baseline Architectures}
\label{secA:baseline_architectures}

\begin{figure*}[t]
    \centering
    \includegraphics[width=0.96\textwidth]{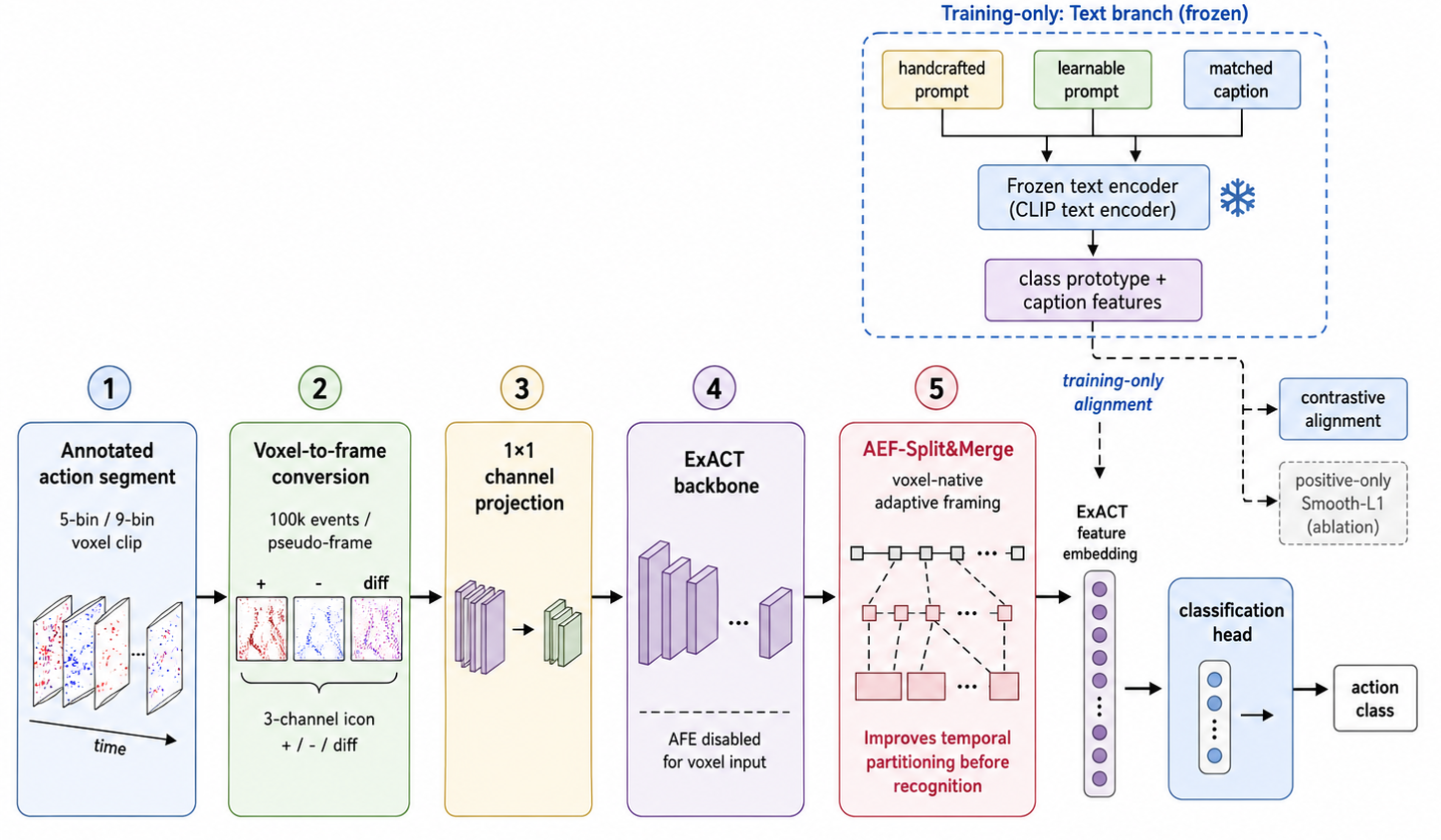}
    \caption{\textbf{ExACT-based recognition adaptation.}
    The diagram distinguishes the original ExACT components from the Event ActivityNet-specific interface. Event ActivityNet voxels pass through AEF-Split{\&}Merge, voxel-to-frame conversion, and an optional $1{\times}1$ projection before the ExACT event encoder. A frozen text encoder processes handcrafted prompts, learnable prompts, and matched ActivityNet captions. Class prototypes and valid captions provide training-time alignment targets; recognition uses the event branch at inference.}
    \label{fig:arch_exact}
\end{figure*}

The benchmark is the primary contribution, and the following diagrams document how published reference architectures are connected to Event ActivityNet. Figures~\ref{fig:arch_exact}--\ref{fig:arch_matr} summarize the adapted ExACT recognition pipeline and the SimOn/MATR causal Online TAL pipelines, respectively. They distinguish inherited modules from our dataset-specific adaptations and do not represent three new model families.

\subsection{ExACT-Based Recognition Adaptation}
\label{secA:arch_exact}

The original ExACT framework first applies Adaptive Fine-grained Event (AFE) representation to a raw event stream, then uses event and frozen text encoders together with its conceptual-reasoning and uncertainty-estimation components~\cite{ZhouZLW24}. Its AFE procedure recursively divides raw event streams according to event-count-image differences and dataset-specific event-count thresholds. Because Event ActivityNet is released as event voxels rather than continuous raw events, we disable the original AFE front end. Our adapted recognition pipeline instead applies voxel-native AEF-Split{\&}Merge, converts the resulting windows into frame-like three-channel polarity tensors, optionally applies a $1{\times}1$ channel projection, and then uses the ExACT event-encoding backbone. The frozen text encoder produces handcrafted and learnable class-prompt embeddings as well as temporally matched caption embeddings. These text features supervise the event encoder through the contrastive or positive-only objectives defined in~\cref{sec:triple_align}; they do not alter the inherited ActivityNet labels. As shown in Figure~\ref{fig:arch_exact}, the Event ActivityNet-specific changes are confined to the voxel interface and training-time text-alignment branch, while the ExACT event encoder remains the recognition backbone.

\subsection{SimOn-Based Online TAL Adaptation}
\label{secA:arch_simon}

\begin{figure*}[!t]
    \centering
    \includegraphics[width=0.96\textwidth]{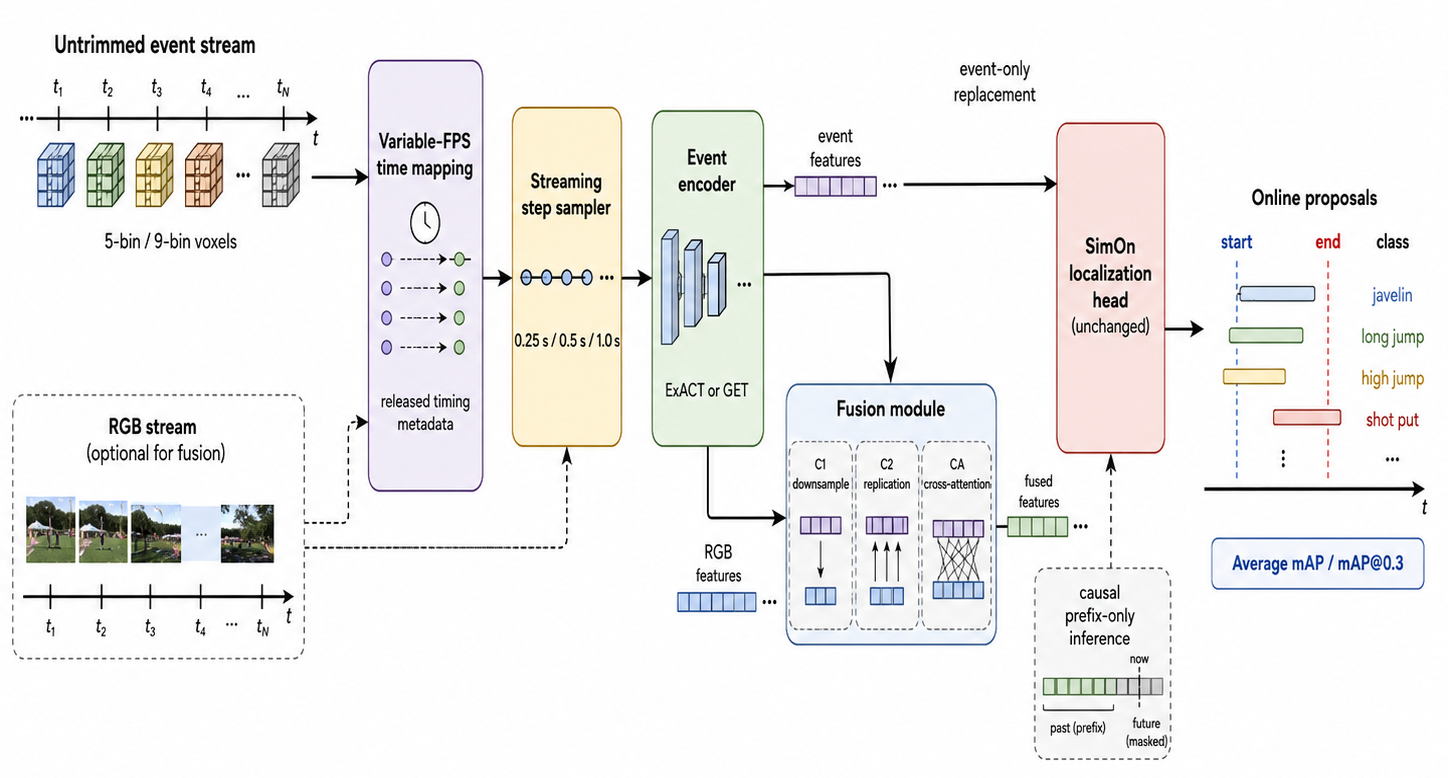}
    \caption{\textbf{SimOn-based causal Online TAL adaptation.}
    The diagram shows RGB-only, event-only, and RGB--event feature construction before the unchanged SimOn head. The matched RGB-only path reuses the fusion model's RGB encoder, aligned timestamps, and input projection without event input. At time $t$, the current feature is the Transformer query, and only cached features and context embeddings from the observed prefix are used as keys and values. Per-class probabilities are converted into online action instances without future access or post-hoc modification.}
    \label{fig:arch_simon}
\end{figure*}

SimOn formulates Online TAL as sequential prediction with a lightweight Transformer~\cite{abs-2211-04905}. At each time step, the current visual feature acts as the query, while recent visual contexts and a learned context embedding form the keys and values. The model emits per-class probabilities, from which action starts and ends are constructed online without access to future frames or retroactive modification. In our adaptation, the SimOn temporal head and causal prediction rule are unchanged. The input is either the matched RGB-only branch, GET/ExACT event features, or one of the aligned RGB--event features (C1, C2, or CA). The RGB-only branch uses the same RGB encoder, temporal alignment, and projection as the fusion model but removes the event stream. The resulting sequence is used as SimOn's current and past visual context, while all caches are updated monotonically. Figure~\ref{fig:arch_simon} makes this separation explicit: modality-specific feature construction occurs before the unchanged causal SimOn head, and every prediction uses only the observed prefix.

\begin{figure*}[t!]
    \centering
    \includegraphics[width=0.96\textwidth]{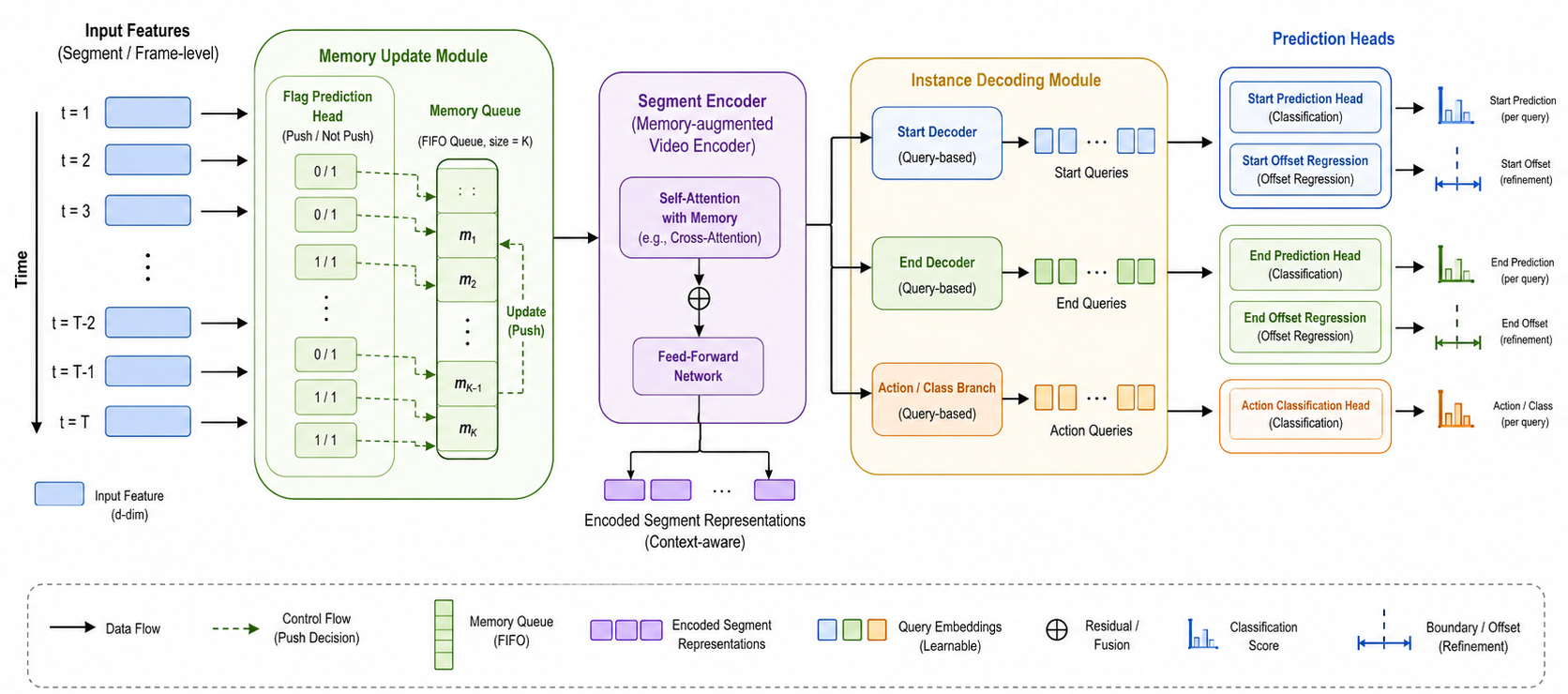}
    \caption{\textbf{MATR-based causal Online TAL adaptation~\cite{song2024online}.}
    The diagram shows event-only or fused segment features entering the original MATR memory-augmented encoder. The end decoder reads the current segment, the start decoder retrieves long-term context from the selective memory queue, and separate class/boundary queries feed the start, end, and action-classification heads. The optional residual heads refine coarse start and end times while retaining the MATR class prediction.}
    \label{fig:arch_matr}
\end{figure*}

\subsection{MATR-Based Online TAL Adaptation}
\label{secA:arch_matr}

MATR contains a feature extractor, a memory-augmented video encoder, an instance decoding module, and separate prediction heads~\cite{song2024online}. A flag-prediction mechanism selectively stores relevant past segment features in a memory queue. The end decoder uses the current encoded segment to locate an action end, whereas the start decoder queries the memory queue to retrieve the corresponding action start. Distinct class and boundary queries decouple action classification from localization. The original start head combines region classification with offset regression, while the end head performs offset regression.

Our adaptation preserves this memory queue, the separate start/end decoding paths, and the causal prediction rule. Event-only or aligned RGB--event features replace the original frame-level input. The C1, C2, and CA settings therefore differ only in feature construction. The offset-head variant additionally applies the residual boundary corrections in~\cref{eq:matr_residual_offsets,eq:matr_refined_proposal} to the coarse MATR proposal. Figure~\ref{fig:arch_matr} shows how the current segment, selective memory queue, and separate class/boundary queries interact, and where the optional residual refinement is attached.

\subsection{Scale-Wise SimOn/MATR Results}
\label{secA:matr_scales}

\Cref{tab:matr_scales} reports the complete Small- and Medium-scale MATR~\cite{song2024online} experiments that underlie the representative comparison in the main paper. The same 5-bin Event ActivityNet protocol is used throughout. C1 downsamples events to the RGB feature rate; C2 replicates RGB features for higher-rate event alignment; CA denotes causal cross-attention; and \texttt{w/ T} fine-tunes the GET event encoder.

\begin{table}[t]
\centering
\caption{\textbf{SimOn and MATR~\cite{song2024online} Online TAL results across Small and Medium.}
Entries are means over three runs. All rows use 5-bin Event ActivityNet data and causal prefix-only inference. SimOn is included as the scale-matched reference. MATR results include C1 concatenation, C2 concatenation, causal cross-attention, and the boundary-refinement variant.}
\label{tab:matr_scales}
\small
\renewcommand{\arraystretch}{1.08}
\resizebox{\columnwidth}{!}{
\begin{tabular}{lccc}
\toprule
\textbf{Scale} & \textbf{Head / Setting} & \textbf{Avg. mAP} & \textbf{mAP@30} \\
\midrule
\multirow{5}{*}{Small}
& SimOn + GET (w/ T) + CA & 25.8 & 40.6 \\
& MATR + GET + C1 & 24.6 & 37.4 \\
& MATR + GET (w/ T) + C2 (1s, 10) & 27.4 & 39.6 \\
& MATR + GET (w/ T) + CA & 28.4 & 41.2 \\
& MATR + GET (w/ T) + CA + offset refinement & \textbf{29.2} & \textbf{44.5} \\
\midrule
\multirow{5}{*}{Medium}
& SimOn + GET (w/ T) + CA & 27.7 & 44.2 \\
& MATR + GET + C1 & 25.2 & 39.4 \\
& MATR + GET (w/ T) + C2 (1s, 10) & 28.2 & 41.6 \\
& MATR + GET (w/ T) + CA & 29.4 & 43.3 \\
& MATR + GET (w/ T) + CA + offset refinement & \textbf{29.9} & \textbf{46.6} \\
\bottomrule
\end{tabular}
}
\end{table}

The scale-wise results show a consistent progression within MATR from C1 to C2 and then to causal cross-attention. On Small, the cross-attention MATR head improves over the scale-matched SimOn reference in both metrics, and boundary refinement further increases mAP@30 from 41.2 to 44.5. On Medium, MATR with cross-attention improves Avg. mAP from 27.7 to 29.4 but is slightly lower at mAP@30 than SimOn (43.3 versus 44.2); the residual boundary-refinement variant raises both metrics to 29.9 and 46.6. Thus, the strongest conclusion is not that every MATR variant dominates SimOn, but that the trend from stronger feature integration to explicit boundary refinement holds at both scales.

\FloatBarrier
\section{AEF-Split{\&}Merge Procedure}
\label{secA:algor}

Algorithm~\ref{alg:AEF} provides the full AEF-Split{\&}Merge procedure
described in Sec.~\ref{sec:AEF}. We start from an initial fixed-length
partition aligned with the voxel window when applicable, greedily apply
splitting until convergence, and then perform greedy merging.

\begin{algorithm}[t]
\caption{AEF-Split{\&}Merge (Greedy)}
\label{alg:AEF}
\small
\begin{algorithmic}[1]
\Require voxel sequence $X$, initial windows $\{W\}$, thresholds $\tau_{\mathrm{split}},\tau_{\mathrm{merge}}$, minimum duration $d_{\min}$
\Ensure adaptive windows $\{W\}$

\Statex \textbf{Split:}
\Repeat
  \State $changed \gets \textbf{false}$
  \ForAll{$W=[t_s,t_e)$ in $\{W\}$}
    \State $t_m^\star \gets \arg\max_{t_m} S_{\mathrm{split}}(W,t_m)$
    \If{$S_{\mathrm{split}}(W,t_m^\star)>\tau_{\mathrm{split}}$ \textbf{and} duration constraints hold}
      \State replace $W$ by $[t_s,t_m^\star)$ and $[t_m^\star,t_e)$
      \State $changed \gets \textbf{true}$
    \EndIf
  \EndFor
\Until{$changed=\textbf{false}$}

\Statex \textbf{Merge:}
\Repeat
  \State $changed \gets \textbf{false}$
  \State find adjacent pair $(W_i,W_{i+1})$ with minimum $S_{\mathrm{merge}}(W_i,W_{i+1})$
  \If{$S_{\mathrm{merge}}(W_i,W_{i+1})<\tau_{\mathrm{merge}}$}
    \State merge them into $W_i\cup W_{i+1}$
    \State $changed \gets \textbf{true}$
  \EndIf
\Until{$changed=\textbf{false}$}
\end{algorithmic}
\end{algorithm}

\FloatBarrier
\section{Limitations}
\label{secA:limit}

Event ActivityNet is derived from conventional RGB videos and therefore
cannot reproduce native sensor noise, high-dynamic-range measurements, or
microsecond timestamps. Voxels follow decoded frame order; mapping ActivityNet
boundaries from seconds to frame/voxel indices uses each video's rational
nominal or average rate and may introduce error for variable-frame-rate
sources. The simulator also resets a deterministic seed per call, so
equal-shaped calls may reuse pseudorandom perturbation patterns. Moreover,
the LPIPS audit measures reconstructable content only at annotated action
centers and does not validate complete temporal boundaries or native-sensor realism. The benchmark is a curated subset of ActivityNet. Class-coverage constraints,
duration stratification, and motion/illumination-oriented enrichment alter the
source distribution, while ActivityNet Captions serve both as an enrichment
signal and optional language supervision. These choices may favor videos whose
captions explicitly describe motion or illumination. Event ActivityNet should
therefore be interpreted as a scalable simulated-event testbed; native-camera
evaluation remains necessary for deployment-oriented claims, and the
multi-terabyte releases may limit accessibility.

\bibliography{aaai2027}

\end{document}